\documentclass[10pt,twocolumn,letterpaper]{article}

\usepackage{iccv}
\usepackage{times}
\usepackage{graphicx}
\usepackage{amsmath}
\usepackage{amssymb}
\usepackage{amsthm}
\usepackage{mathrsfs}
\usepackage{bbm}
\usepackage{subfigure}
\usepackage{multirow}

\usepackage[pagebackref=true,breaklinks=true,letterpaper=true,colorlinks,bookmarks=false]{hyperref}

\iccvfinalcopy 

\ificcvfinal\fi

\begin{document}

\title{Structure First Detail Next: Image Inpainting with Pyramid Generator}

\author{Shuyi Qu$^{1,2}$\quad Zhenxing Niu$^{2}$\quad Kaizhu Huang$^{1}$\quad Jianke Zhu$^{3}$\quad \\Matan Protter$^{2}$\quad Gadi Zimerman$^{2}$\quad Yinghui Xu$^{2}$\\
$^{1}$Xi’an Jiaotong-Liverpool University\quad $^{2}$Alibaba DAMO Academy\quad\\ $^{3}$Zhejiang University\\
\tt\small  $^1${\tt\small{firstname.lastname@xjtlu.edu.cn}}
 \quad
 $^2${\tt\small{firstname.lastname@alibaba-inc.com}}
  \quad\\
 $^3${\tt\small{jkzhu@zju.edu.cn}}
 }

\maketitle
\begin{abstract}
Recent deep generative models have achieved promising performance in image inpainting. However, it is still very challenging for a neural network to generate realistic image details and textures, due to its inherent spectral bias. By our understanding of how artists work, we suggest to adopt a `structure first detail next' workflow for image inpainting. To this end, we propose to build a Pyramid Generator by stacking several sub-generators, where lower-layer sub-generators focus on restoring image structures while the higher-layer sub-generators emphasize image details. Given an input image, it will be gradually restored by going through the entire pyramid in a bottom-up fashion. Particularly, our approach has a learning scheme of progressively increasing hole size, which allows it to restore large-hole images. In addition, our method could fully exploit the benefits of learning with high-resolution images, and hence is suitable for high-resolution image inpainting. Extensive experimental results on benchmark datasets have validated the effectiveness of our approach compared with state-of-the-art methods.
\end{abstract}


\section{Introduction}

Image inpainting aims at restoring corrupted images with reasonable and relevant contents, which is widely used in many real-world applications including image restoration, object removal, and photo editing. In recent years, deep generative models are broadly adopted to solve this problem, and have achieved impressive results~\cite{li2017generative,yan2018shift,yu2018generative,liu2019coherent,zheng2019pluralistic,ren2019structureflow,yu2020region,li2020recurrent,liao2020guidance,Liu2019MEDFE,zeng2020high}. However, it is still very challenging to generate coherent and realistic image details \cite{yu2019free,Soo2020Zoom}. Such challenges probably stem from the neural networks' spectral bias \cite{Nasim2019}, \emph{i.e.}, neural networks are biased towards learning low frequency components instead of high frequency details. To alleviate this problem, most state-of-the-art methods adopt a coarse-to-fine framework. Taking the DeepFillv2 \cite{yu2019free} as an example, images are first restored with an coarse network, and then the details are further refined at the next stage.

Furthermore, there is a conflict in modeling image global structures and image local details~\cite{HR17,Soo2020Zoom}. To tackle this problem, Yang \emph{et al}. propose a joint loss function where the content term and texture term are used to model image global structure and local texture respectively~\cite{HR17}. However, modeling image structures often prefers to learn from large image regions or even the whole image, while modeling image details prefers to learn from small image patches. This dilemma requires different receptive field sizes, and it is hard to satisfy in single neural network.

By our understanding, image inpainting looks like artists drawing a picture for a scene. Artists usually first draw the global structures/sketches of the scene and then refine its local details \cite{Edwards2012,Eitz2012,nazeri2019edgeconnect}. Inspired by that, we suggest to explicitly separate the restoration of image structures from that of details, and follow the philosophy of `\textbf{structure first detail next}'. Thus, we propose to engage distinct sub-generators to restore image structures and details, respectively. 

In particular, we build a Pyramid Generator for image inpainting. The pyramid generator is constructed by stacking several sub-generators from bottom to top layer. Inputs are downsampled into different resolutions and are fed into corresponding sub-generators. The sub-generators at low-layers focus on restoring the image global structures, while the high-layer sub-generators focus on restoring image local details. Given an image, it will be gradually inpainted by going through the pyramid from bottom to top. Obviously, our novel inpainting approach actually takes the `structure first detail next' workflow.

Note that the previous \emph{coarse-to-fine frameworks} in~\cite{yu2019free} only cascade two similar sub-generators (\emph{i.e.}, they are trained with inputs of same resolution) and fail to separate effectively the image structures restoration and image details restoration. On the contrary, with the \emph{multi-scale multi-layer stacking framework}, our pyramid generator can not only effectively separate the structures and details modeling and indeed practise restoring structures before details.

It is well-known that image inpainting becomes much more challenging when the corrupted area is relative large \cite{liu2018image,Soo2020Zoom,HiFill}. Some recent work \cite{Soo2020Zoom} illustrate that such \textbf{large-hole challenges} could be effectively alleviated by a progressive learning strategy, \emph{i.e.}, first learn to restore small hole and then learn to restore a large hole. 

\begin{figure}
\centerline{\includegraphics[width=0.4\textwidth]{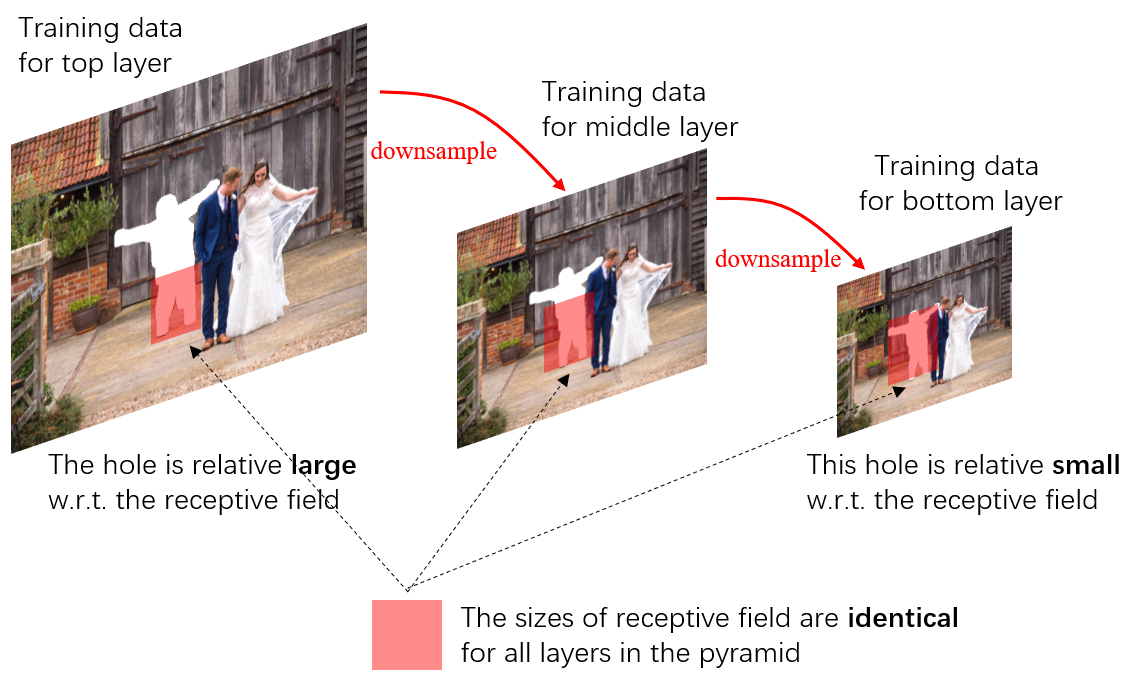}}
\caption{Bottom-layer inpainting task (right) looks like a small-hole inpainting task, while the top-layer inpainting (left) looks like a large-hole inpainting. It is obvious that the mask size of bottom layer is relative small with respect to the same receptive field size, while the mask size of top layer is relative large.}
\label{fig:large_mask}
\end{figure}

Our approach exactly aligns to such strategy. As shown in Figure~\ref{fig:model}, both input image and mask image are simultaneously downsampled in our pyramid, thus the ratio of hole size to image size is identical for all layers. However, whether the hole size is large or small should be measured with respect to the size of convolutional receptive field (instead of image size). As shown in Figure~\ref{fig:large_mask}, since all sub-generators at different layers have the same architecture as well as the same receptive field size, the hole size is \emph{relatively} small for the low-layer sub-generator. Thus, the low-layer inpainting looks like the small-hole inpainting task. In contrast, the high-layer inpainting looks like large-hole inpainting. With our bottom-up inpainting workflow, we are actually aligning to the hole-increasing strategy, \emph{i.e.}, first conduct small-hole inpainting and then large-hole inpainting. As a result, our pyramid generator has advantages of dealing with large holes.

Finally, our pyramid generator is more suitable for \textbf{high-resolution image inpainting}. Many previous works have shown that increasing the resolution of training images could benefit the high-resolution image inpainting \cite{zeng2019learning,HiFill}. Our experimental results also validate such observation, as shown in Section~\ref{section:reso}. However, we find that such performance gain is still limited if we just directly train the existing models (\emph{e.g.}, DeepFillv2 \cite{yu2019free}) with high-resolution images (\emph{e.g.}, on 512$\times$512 images), as shown in Fig~\ref{fig:reso}. 

We argue that such observation is also related to the large-hole challenge. Since high-resolution training images often come along with large holes, if we directly train a generator with high-resolution images, we have to face the large-hole challenge. Nevertheless, due to the capability of handling the large-hole challenge, our approach could fully exploit the advantages of learning with high-resolution images. As a result, our approach is more suitable for high-resolution image inpainting.

To the best of our knowledge, this is the first work that adopts a pyramid of GANs for image inpainting. We highlight our contributions as follows:
\begin{itemize}
\item A Pyramid Generator is proposed to conduct image inpainting by following the strategy of `structure first detail next'. With our dedicated multi-scale generative architecture, the conflict between image global structures and local details restoration could be alleviated.
\item Our pyramid generator has a learning scheme of progressively increasing hole size, which allows it to restore large holes.
\item Our approach could totally reap the benefits of learning with high-resolution images, and hence is suitable for inpainting high-resolution images.
\end{itemize}

\section{Related Work}
\subsection{Image Inpainting}
Conventional image inpainting methods often utilize low-level image statistics~\cite{ballester2001filling}. PatchMatch \cite{barnes2009patchmatch} fills holes by searching similar patches from unfilled area. Although effective at textured images, when inpainting complex images with global structures, these methods often generate artifacts or incoherent content.

Recently, many deep learning based methods are proposed with great improvement~\cite{pathak2016context,iizuka2017globally,liu2018image,yu2019free,HiFill}. Context encoder \cite{pathak2016context} presents the first attempt to apply the convolution neural network on image inpainting. Iizuka \etal improve the architecture with both global and local discriminators to keep consistency \cite{iizuka2017globally}. Partial convolution \cite{liu2018image} is proposed to handle free-formed masks by using only valid pixels as conducting convolutions. Recently, DeepFillv2 \cite{yu2019free} introduces a contextual attention module and a coarse-to-fine learning framework, which has significantly improved the inpainting performance. HiFill \cite{HiFill} focuses on high-resolution inpainting task and proposes a contextual residual aggregation mechanism.

\begin{figure*}\vspace{-0.3cm}
\centerline{\includegraphics[width=0.9\textwidth]{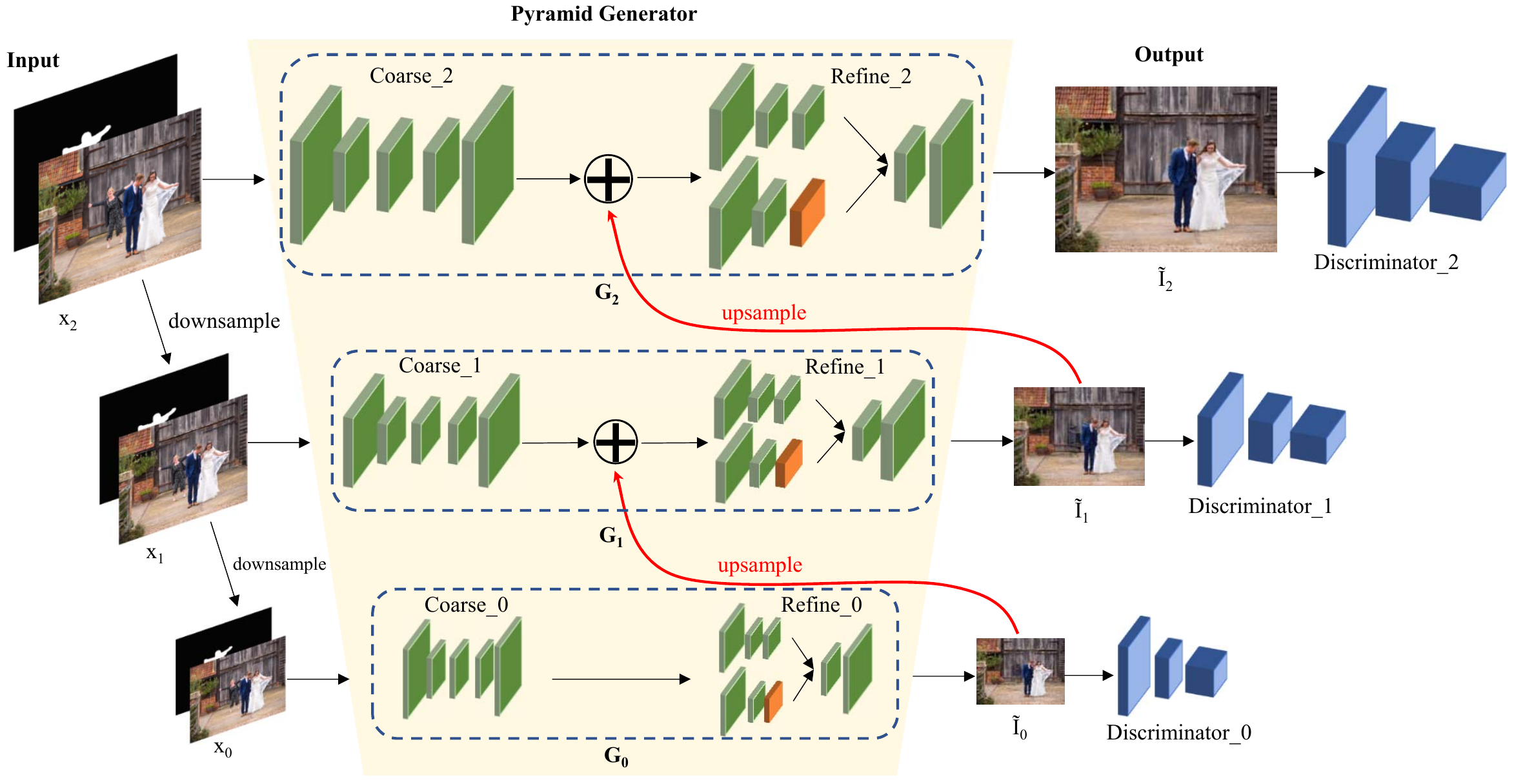}}
\caption{Architecture of our pyramid generator. The input image and mask are gradually downsampled until to the bottom layer, and then the model is trained in a bottom-up manner. The output of a lower layer will be fed to its higher layer to help refining the higher-layer's results. $\tilde{I}_{2}$ is regarded as the final output of our pyramid generator.}
\label{fig:model}
\end{figure*}

\subsection{Multi-scale Mechanism}
Multi-scale design is applied broadly in many computer vision tasks thanks to its progressive refinement property \cite{eigen2015predicting,ren2016single,wang2018high,shaham2019singan}. One representative work is SinGAN \cite{shaham2019singan}, where a pyramid of GANs is proposed to generate a realistic image sample of arbitrary size and aspect ratio. In particular, the pyramid GANs has a multi-scale framework: images are gradually generated with the increasing of image scale, from bottom layer to the top one.

Multi-scale mechanism is also adopted to image inpainting task. 
In \cite{HR17}, images are gradually refined by increasing image scales, but only $L2$ content loss, VGG textual loss and TV loss are used for training. Since adversarial loss is not adopted for learning, it is still hard to generate realistic image details. 
Recently, a pyramid-context encoder PEN is proposed to incorporate high-level semantics with low-level pixels in \cite{zeng2019learning}. PEN has only one encoder, and there is a feature-pyramid within the encoder.

On the contrary, our model is similar to SinGAN that has a stack of distinct sub-generators, where all the sub-generators are trained in an adversarial manner with independent discriminators. We argue that those distinct sub-generators could better model image structures and local details respectively.


\section{Method}\label{section:Approach}
In this section, we first introduce the architecture of our pyramid generator including its layer fusion strategy and the adaptive dilation scheme. Next, we discuss the loss function and describe how to effectively train our pyramid generator.

\subsection{Pyramid Generator}
In order to fully capture the internal statistics of an image at different scales, \cite{shaham2019singan} proposes a hierarchical GAN architecture to learn the distribution within each scale separately. Inspired by that, we adopt its hierarchical framework to separate the modeling of image global structures and local details at different scales. 

\subsubsection{Architecture Design}\label{section:pg} Figure~\ref{fig:model} shows the architecture of our pyramid generator which are stacked from some sub-generators. DeepFillv2 \cite{yu2019free} is adopted as the sub-generator in this paper due to its state-of-the-art performance. At different layers, the sub-generator is trained with images of different resolutions. 

Specifically, the pyramid consists of $N+1$ sub-generators \{$G_{0}$, ..., $G_{N}$\}. Their inputs are \{$x_{0}$, ..., $x_{N}$\}, where each $x_{n}$ is a concatenation of an image and a mask. In this paper, we adopt a three-layer architecture (\emph{i.e.}, $N=2$) since it is enough to inpaint high-resolution images. 

We train the model on 512$\times$512 images, so the input image is of 512$\times$512 resolution. The mask is a binary image, where value of 1 represents corrupted area and 0 represents known area. As shown in Figure~\ref{fig:model}, the original training image and mask are noted as $x_{2}$ (with resolution of 512$\times$512), which are gradually downsampled to $x_{1}$ (with resolution of 256$\times$256) and $x_{0}$ (with resolution of 128$\times$128), respectively.

In the pyramid, each layer has a sub-generator $G_{n}$ and a corresponding discriminator $D_{n}$. The model is trained in a bottom-up manner: 1) $G_{0}$ is trained firstly by using $x_{0}$. Let $\tilde{I_{n}}$ indicate the recovered image of each layer. Thus, we have the output of bottom layer,
\begin{equation}
\tilde{I_{0}} = G_{0}(x_{0}).\label{eq1}
\end{equation}
2) The output $\tilde{I}_{0}$ is then upsampled and fed to sub-generator $G_{1}$, meanwhile  $x_{1}$ is also fed to $G_{1}$. Both of them are used to train $G_{1}$. The training of $G_{2}$ is  the same as $G_{1}$. As a result, we have the following outputs,
\begin{equation}
\tilde{I}_{n} = G_{n}(x_{n}, \tilde{I}_{n-1}), n>0.\label{eq2}
\end{equation}
Note that  $\tilde{I}_{2}$ is regarded as the final output of our pyramid generator.

\subsubsection{Fusion Strategy} 
In our pyramid generator, the output of a lower layer will be fed to a higher layer to help refine the higher-layer's results. For example, the output of $G_{0}$ will be fed to $G_{1}$ to refine the results of $G_{1}$. There are many possible fusion choices. Since there are two stages for each $G_{n}$ (\emph{i.e.}, coarse-stage and refine-stage), the output of $G_{0}$ could be fed to either the coarse-stage or the refine-stage of $G_{1}$.

On the other hand, the fusion could be conducted in feature-level or image-level. Feature-level fusion is similar to \cite{wang2018high} that the feature maps of $G_{0}$ is upsampled and added to the corresponding feature maps of $G_{1}$. In contrast, the image-level fusion is to upsample the output image of $G_{0}$ and add to the internal produced image of $G_{1}$.

We have conducted extensive experiments to compare all those fusion options, and selected the best one according to experimental results. Specifically, we adopt the refine-stage as well as image-level fusion strategy. As shown in Figure~\ref{fig:model}, the output image of $G_{0}$ is upsampled and added with the output of coarse stage of $G_{1}$. Then their sum is fed to the refine stage of $G_{1}$. This process can be formulated as
\begin{equation}
\tilde{I}_{n} = G_{n}^{r}(G_{n}^{c}(x_{n}) + upsample(\tilde{I}_{n-1})), n>0,\label{eq3}
\end{equation}
where $G_{n}^{c}$ and $G_{n}^{r}$ indicate the coarse and refine stage of $G_{n}$ respectively. 

The detailed comparisons of those options are shown in Section~\ref{section:ablation}. We guess that refine-stage fusion being better than coarse-stage fusion is due to that the high-frequency  details of high-resolution image (\emph{e.g.}, $x_{1}$) can be sufficiently exploited by the $G_{1}^{c}$ before conducting fusion. By receiving both the output of $G_{1}^{c}$ and $G_{0}$, the refine network $G_{1}^{r}$ could properly leverage the good results of image details modeling (by $G_{1}^{c}$) and image structure modeling (by $G_{0}$). 

Regarding to the feature-level fusion, we found that it is either inferior to image-level fusion or relatively difficult to stably achieve convergence. Besides, we also try another option that is to directly drop the coarse-stage $G_{1}^{c}$ and only remain refine-stage $G_{1}^{r}$. Although it can reduce some computation, it is also difficult to stably get convergence.

\subsubsection{Adaptive Dilation}
Dilated convolutions~\cite{iizuka2017globally} are broadly used in inpainting neural network, since it can not only explicitly adjust filter’s field-of-view but also keep the resolution of feature maps. Keeping the resolution of feature maps is very important to the performance of those pixel-level algorithms such as image inpainting, and image segmentation. 

On the other hand, adjusting filter’s field-of-view is conducted by adjusting the \emph{dilation rate} of dilated convolutions, which will determine the receptive field of convolutions. However, there is a conflict in modeling image global structures and image local details: modeling image structures often prefers to observe large image regions, while modeling image details prefers to observe small image patches. Therefore, it is very hard to select a proper \emph{dilation rate} for a single-scale architecture to balance the distinct needs of modeling image structures and details.

Nevertheless, our pyramid generator has several independent sub-generators so that we could select distinct dilation rate for different sub-generators. Specifically, $G_{0}$ emphasizes modeling global structures and needs a relatively large receptive field, and hence we have 4 dilated convs with dilation rates $\{2, 4, 8, 12\}$. In contrast, $G_{1}$ and $G_{2}$ emphasize modeling image local details, and hence we only have 3 dilated convs with the dilation rates $\{2, 4, 8\}$. In this way, we can effectively solve the conflict of modeling image global structures and local details, which is called as adaptive dilation scheme in this paper.

\begin{figure*}[htbp]
    \centering
    \subfigure[Masked]{
    \begin{minipage}[b]{0.11\linewidth}
    \includegraphics[width=1\linewidth]{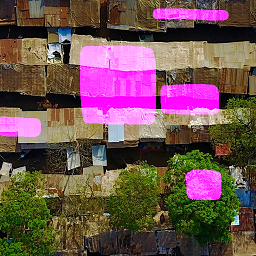}\vspace{1pt}
    \includegraphics[width=1\linewidth]{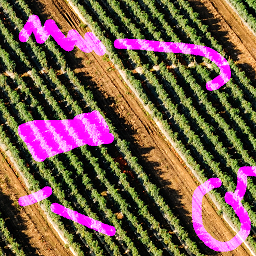}\vspace{1pt}
    \includegraphics[width=1\linewidth]{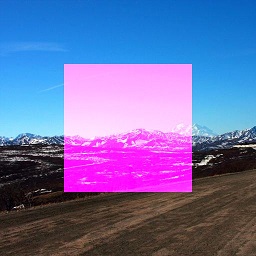}\vspace{1pt}
    \includegraphics[width=1\linewidth]{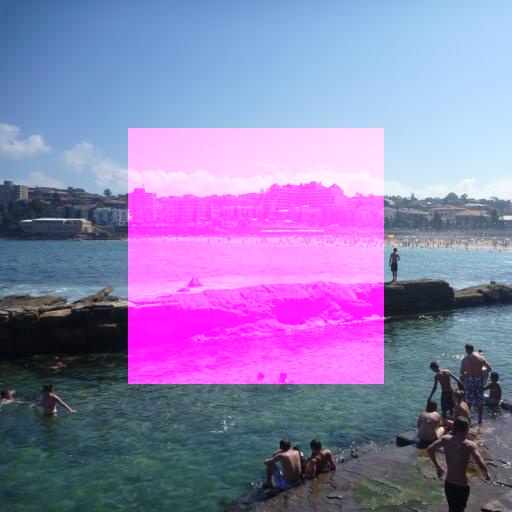}\vspace{1pt}
    \includegraphics[width=1\linewidth]{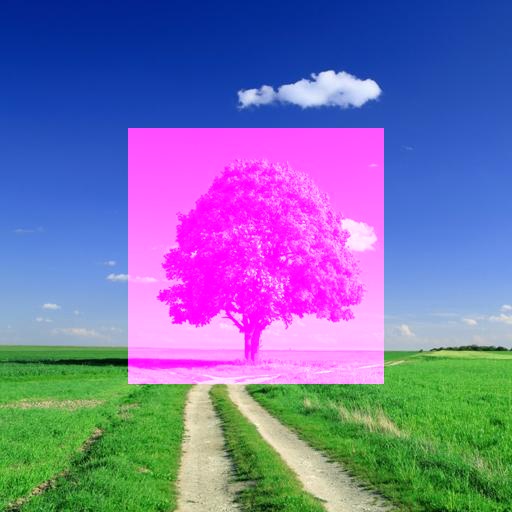}\vspace{1pt}
    \end{minipage}}
    \subfigure[G\&L]{
    \begin{minipage}[b]{0.11\linewidth}
    \includegraphics[width=1\linewidth]{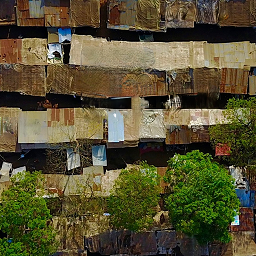}\vspace{1pt}
    \includegraphics[width=1\linewidth]{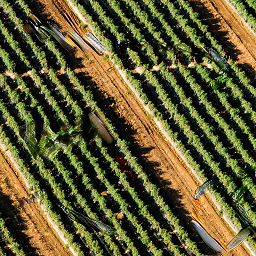}\vspace{1pt}
    \includegraphics[width=1\linewidth]{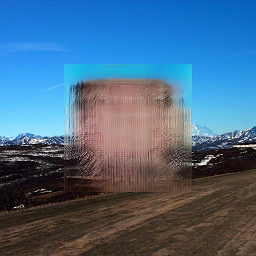}\vspace{1pt}
    \includegraphics[width=1\linewidth]{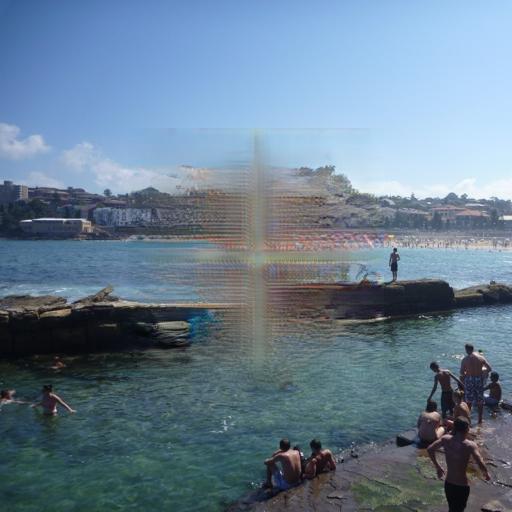}\vspace{1pt}
    \includegraphics[width=1\linewidth]{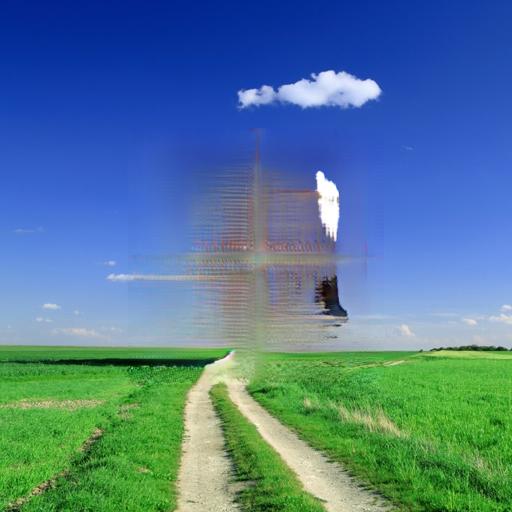}\vspace{1pt}
    \end{minipage}}
    \subfigure[EC]{
    \begin{minipage}[b]{0.11\linewidth}
    \includegraphics[width=1\linewidth]{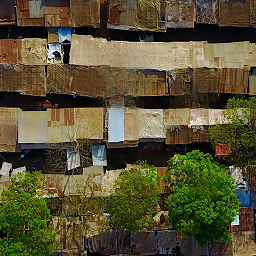}\vspace{1pt}
    \includegraphics[width=1\linewidth]{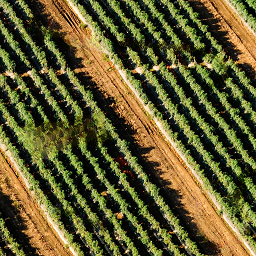}\vspace{1pt}
    \includegraphics[width=1\linewidth]{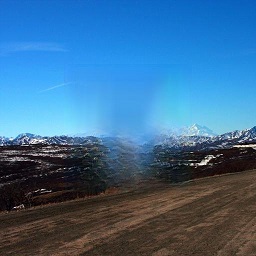}\vspace{1pt}
    \includegraphics[width=1\linewidth]{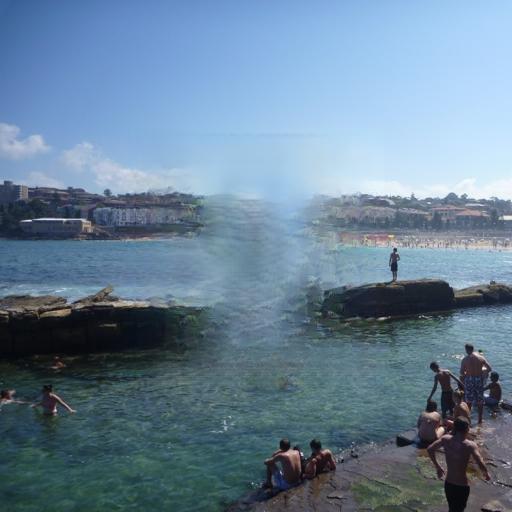}\vspace{1pt}
    \includegraphics[width=1\linewidth]{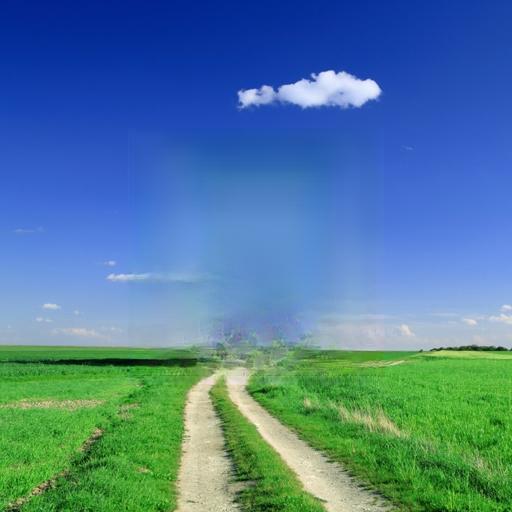}\vspace{1pt}
    \end{minipage}}
    \subfigure[PEN]{
    \begin{minipage}[b]{0.11\linewidth}
    \includegraphics[width=1\linewidth]{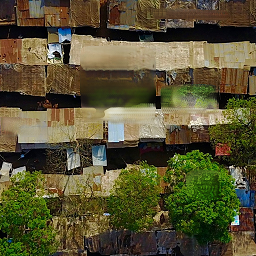}\vspace{1pt}
    \includegraphics[width=1\linewidth]{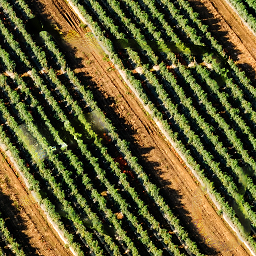}\vspace{1pt}
    \includegraphics[width=1\linewidth]{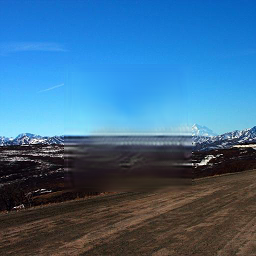}\vspace{1pt}
    \includegraphics[width=1\linewidth]{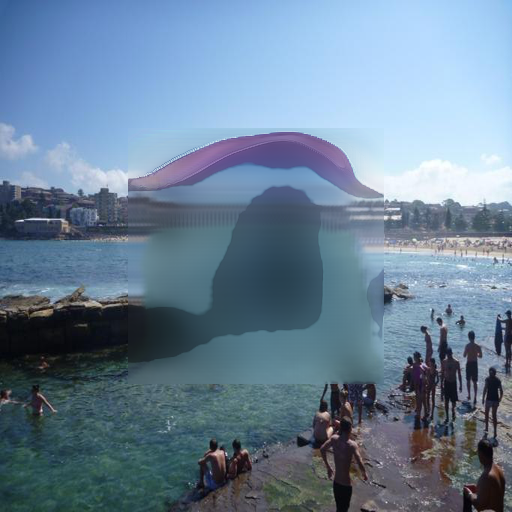}\vspace{1pt}
    \includegraphics[width=1\linewidth]{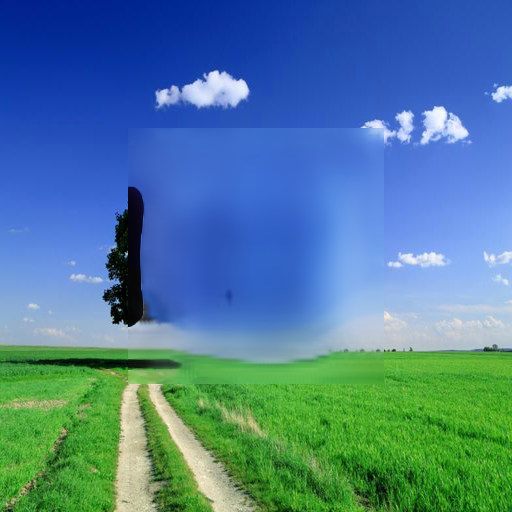}\vspace{1pt}
    \end{minipage}}
    \subfigure[DeepFillv2]{
    \begin{minipage}[b]{0.11\linewidth}
    \includegraphics[width=1\linewidth]{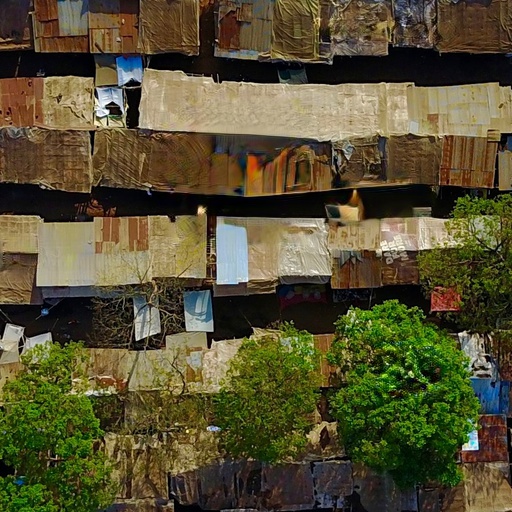}\vspace{1pt}
    \includegraphics[width=1\linewidth]{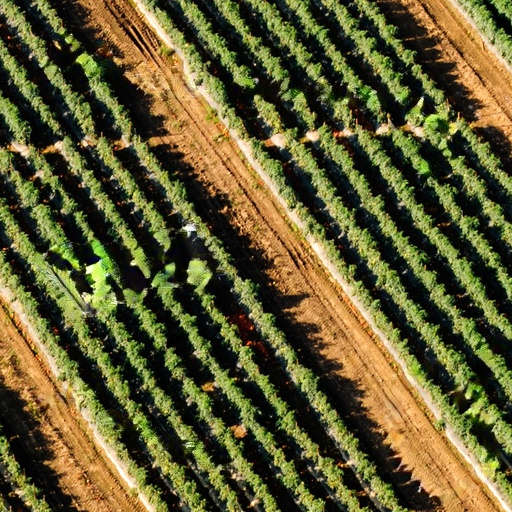}\vspace{1pt}
    \includegraphics[width=1\linewidth]{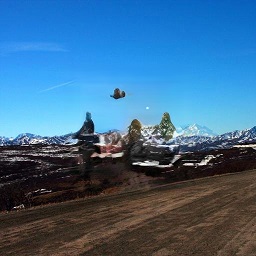}\vspace{1pt}
    \includegraphics[width=1\linewidth]{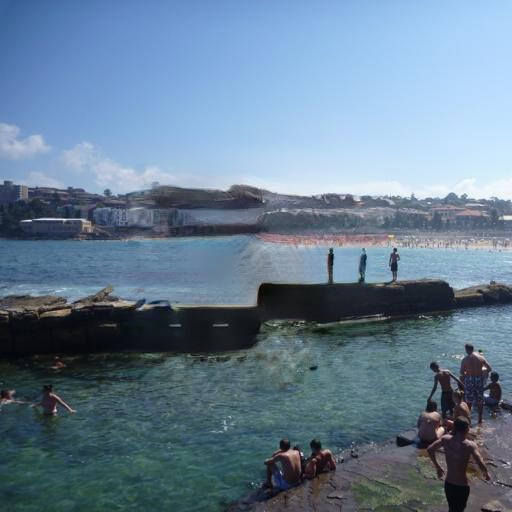}\vspace{1pt}
    \includegraphics[width=1\linewidth]{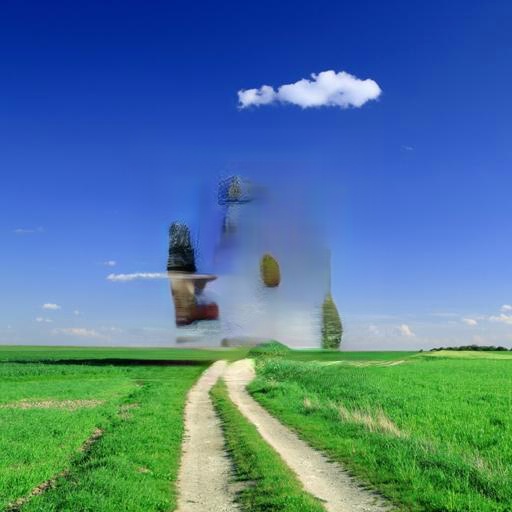}\vspace{1pt}
    \end{minipage}}
    \subfigure[DeepFillv2$^\dagger$]{
    \begin{minipage}[b]{0.11\linewidth}
    \includegraphics[width=1\linewidth]{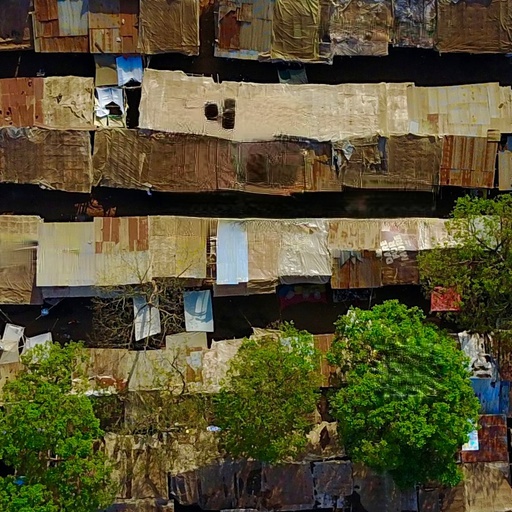}\vspace{1pt}
    \includegraphics[width=1\linewidth]{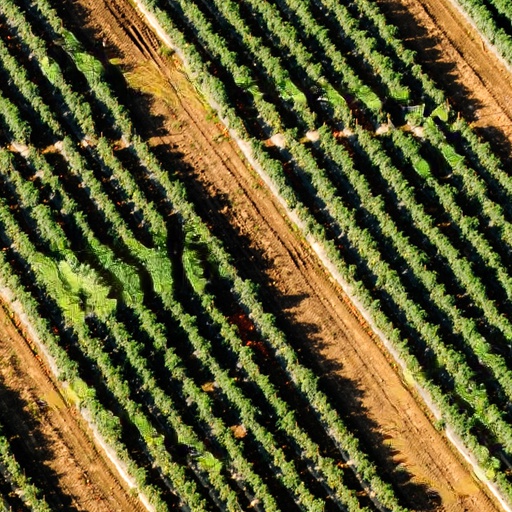}\vspace{1pt}
    \includegraphics[width=1\linewidth]{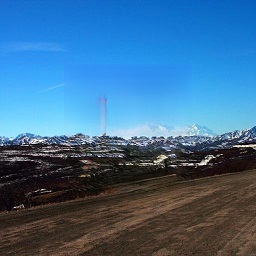}\vspace{1pt}
    \includegraphics[width=1\linewidth]{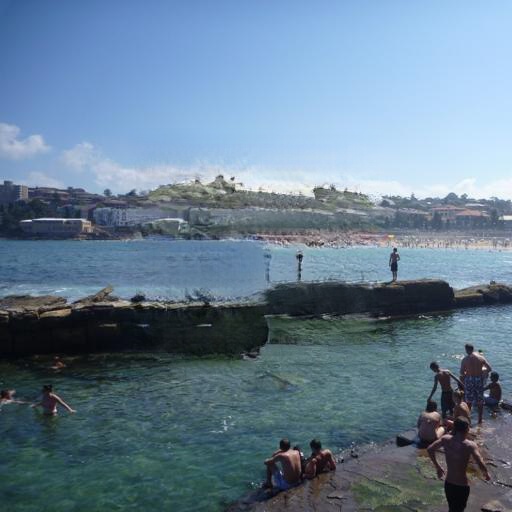}\vspace{1pt}
    \includegraphics[width=1\linewidth]{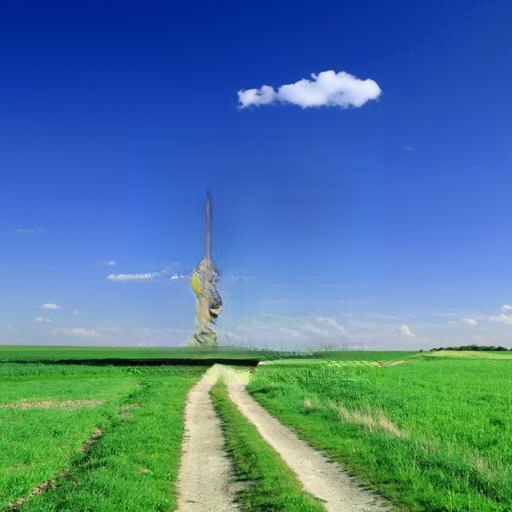}\vspace{1pt}
    \end{minipage}}
    \subfigure[HiFill]{
    \begin{minipage}[b]{0.11\linewidth}
    \includegraphics[width=1\linewidth]{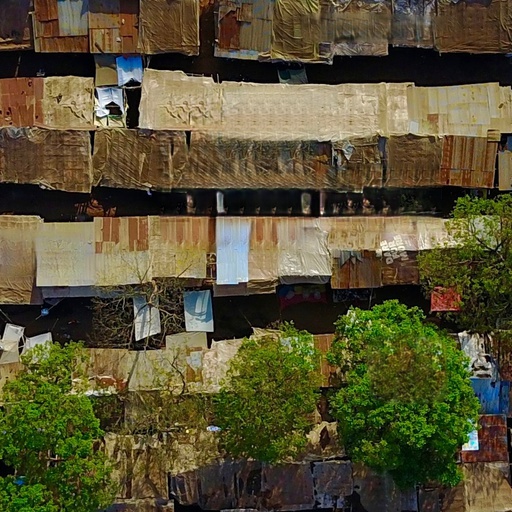}\vspace{1pt}
    \includegraphics[width=1\linewidth]{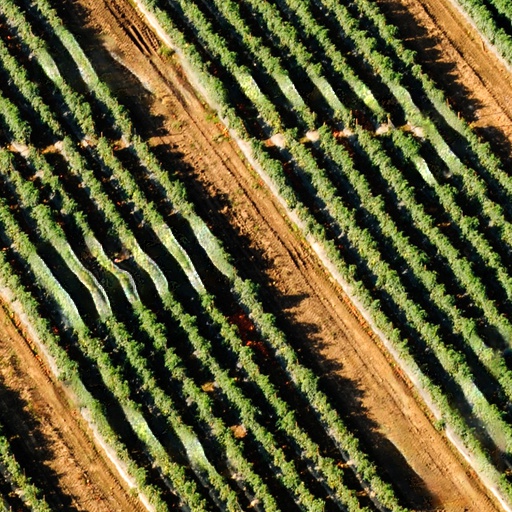}\vspace{1pt}
    \includegraphics[width=1\linewidth]{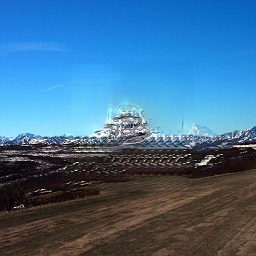}\vspace{1pt}
    \includegraphics[width=1\linewidth]{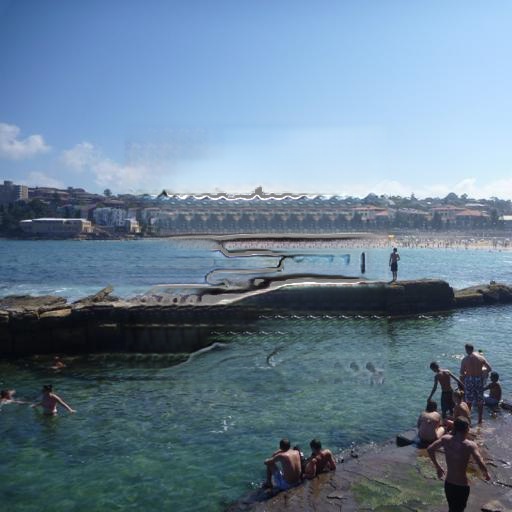}\vspace{1pt}
    \includegraphics[width=1\linewidth]{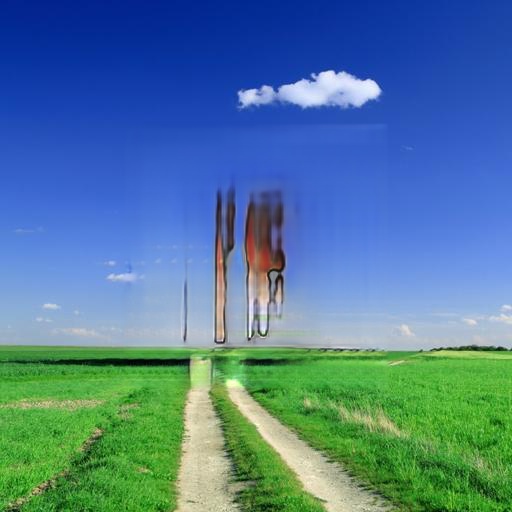}\vspace{1pt}
    \end{minipage}}
    \subfigure[Ours]{
    \begin{minipage}[b]{0.11\linewidth}
    \includegraphics[width=1\linewidth]{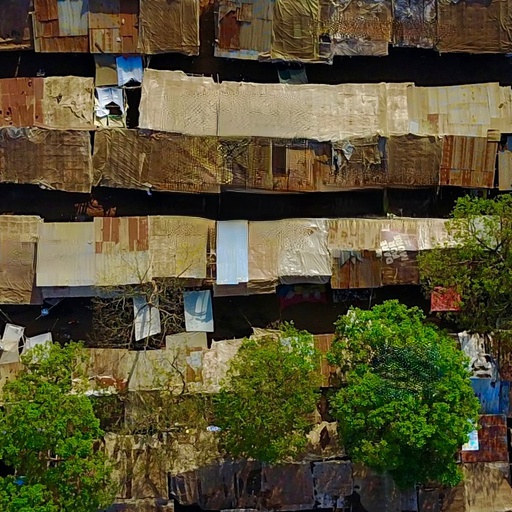}\vspace{1pt}
    \includegraphics[width=1\linewidth]{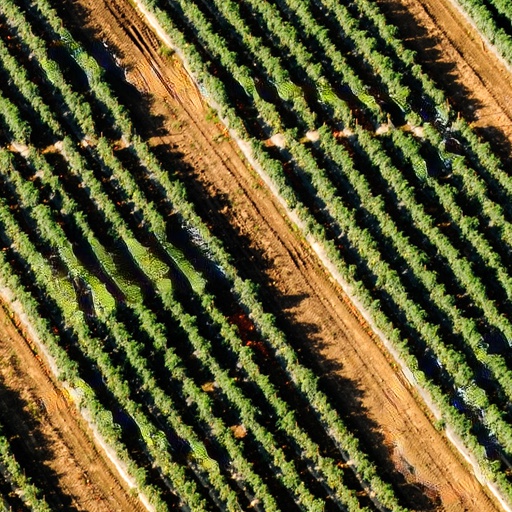}\vspace{1pt}
    \includegraphics[width=1\linewidth]{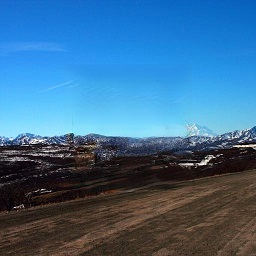}\vspace{1pt}
    \includegraphics[width=1\linewidth]{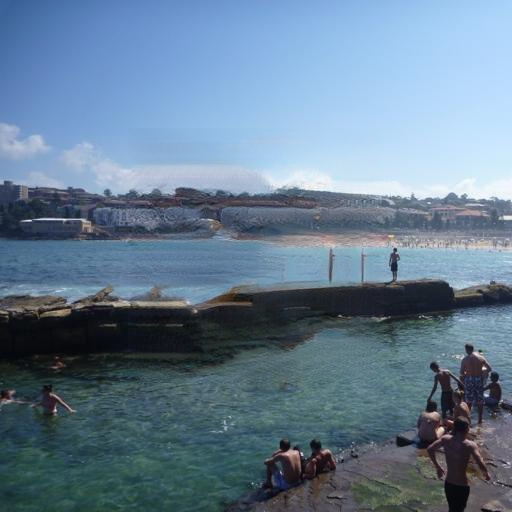}\vspace{1pt}
    \includegraphics[width=1\linewidth]{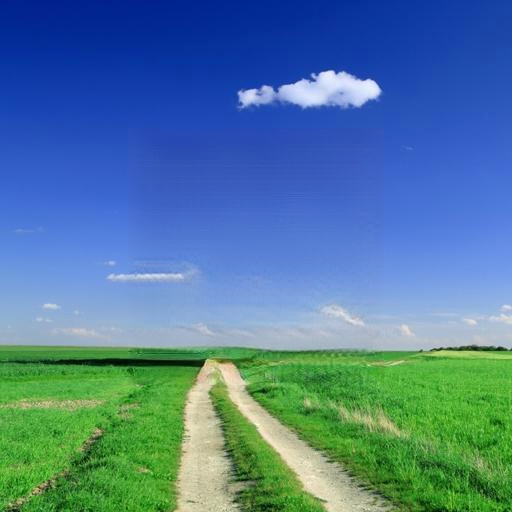}\vspace{1pt}
    \end{minipage}}
    \caption{Qualitative comparisons on Places2 val and DIV2K val set (with image resolution 512$\times$512). $^\dagger$ indicates methods retrained by us with 512$\times$512 images. Best viewed by zooming-in.}
    \label{fig:visual}
\end{figure*}

\begin{table*}[htbp]
\scriptsize
\centering
\caption{Quantitative comparisons on the Places2 val and DIV2K val set (with image resolution 512$\times$512). Both center masks and free-form masks settings are considered. Up-arrow ($\uparrow$) indicates higher score is better, while lower score is better for down-arrow ($\downarrow$). $^\dagger$ indicates that the model is trained on 512$\times$512 images instead of 256$\times$256 images. $^*$: This is reported from \cite{HR17} using single Titan X GPU.}
\begin{tabular}{c|ccc|ccc|ccc|ccc|c}
\hline
\multirow{2}*{Method}&\multicolumn{3}{c|}{Places2 square mask}& \multicolumn{3}{c|}{Places2 free-form mask}&\multicolumn{3}{c|}{DIV2K square mask}&\multicolumn{3}{c|}{DIV2K free-form mask}&speed\\
&SSIM$\uparrow$&PSNR$\uparrow$&L1$\downarrow$&SSIM$\uparrow$&PSNR$\uparrow$&L1$\downarrow$&SSIM$\uparrow$&PSNR$\uparrow$&L1$\downarrow$&SSIM$\uparrow$&PSNR$\uparrow$&L1$\downarrow$&time$/$image\\
\hline
GL \cite{iizuka2017globally}& 0.773& \textbf{20.72}& 0.055&0.683& 18.71& 0.086&0.769&20.85&0.075&0.660&17.88&0.109&125ms\\
MNPS \cite{HR17}& 0.727& 19.22& 0.058& -& - &- & 0.749&23.15&0.047&- & -&-&1min$^*$\\
EC \cite{nazeri2019edgeconnect}&0.770&19.55&0.069&0.736&18.92&0.086&0.817&24.65&0.042&0.759&21.61&0.070&178ms\\
PEN \cite{zeng2019learning}& 0.754& 19.36& 0.056&0.724& 19.72& 0.068&0.792&23.42&0.036&0.758&22.40&0.053&347ms\\
DeepFillv2 \cite{yu2019free}& 0.769& 18.86& 0.056& 0.737& 18.59& 0.073&0.814&23.47&0.043&0.765&20.91&0.064&69ms\\
DeepFillv2$^\dagger$ \cite{yu2019free}& 0.769& 19.40& 0.053& 0.740& 19.41& 0.067&0.827&25.72&0.036&0.789&23.40&0.053&69ms\\
HiFill \cite{HiFill}& 0.749& 19.87& 0.051&0.701& 19.14& 0.074&0.791&23.63&0.040&0.736&20.67&0.073&24ms\\
\hline
\textbf{ours}& \textbf{0.777}& 20.09&\textbf{0.050}& \textbf{0.746}& \textbf{19.90}& \textbf{0.064}&\textbf{0.840}&\textbf{26.74}&\textbf{0.034}&\textbf{0.796}&\textbf{24.25}&\textbf{0.050}&85ms\\
\end{tabular}
\label{metric}
\end{table*}

\subsection{Learning of Pyramid Generator}
In this section, we will describe the loss function and the training details of our pyramid generator. Our pyramid generator is composed of several sub-generators, where each sub-generator is trained by using GAN mechanism. Particularly, for each sub-generator $G_{n}$ we adopt the loss function as \cite{yu2019free}, which consists of a reconstruction term and an adversarial term,
\begin{equation}
\mathcal{L}_{n} = \mathcal{L}_{adv}(G_{n}, D_{n})+\alpha \mathcal{L}_{re}(G_{n}),\label{eq4}
\end{equation}
and we choose $\alpha=1$ in our experiments. 

The reconstruction term is defined as the $L1$ distance between the generated output $\tilde{I}_{n}$ and the ground-truth image $I_{n}$ at the pixel level:
\begin{equation}
\mathcal{L}_{re}(G_{n}) = ||\tilde{I}_{n}-I_{n}||_{1}.\label{eq5}
\end{equation}

For the adversarial term, we use the hinge loss. The loss for generator is
\begin{equation}
\mathcal{L}_{G} = -\mathbb{E}_{z\sim{p_{z}}, y\sim{p_{data}}}D\left(G\left(z\right), y\right),\label{eq6}
\end{equation}
and for discriminator, it is
\begin{equation}   
\begin{split}
\mathcal{L}_{D} = \mathbb{E}_{x\sim{p_{data(x)}}}ReLU\left(\mathbbm{1}-D\left(x\right)\right)\\
+\mathbb{E}_{z\sim{p_{z(z)}}}ReLU\left(\mathbbm{1}+D\left(G\left(z\right)\right)\right).\label{eq7}
\end{split}
\end{equation}

Taking the three-layer pyramid generator as an example, its total loss can be represented by
\begin{equation}
\mathcal{L}_{PG} = \lambda_{0}\mathcal{L}_{0}+\lambda_{1}\mathcal{L}_{1}+\lambda_{2}\mathcal{L}_{2},\label{eq8}
\end{equation}
where $\mathcal{L}_{0}$, $\mathcal{L}_{1}$ and $\mathcal{L}_{2}$ indicate the loss of $G_{0}$, $G_{1}$, and $G_{2}$ respectively, with $\lambda_{0}$, $\lambda_{1}$, $\lambda_{2}$ the weights for them. In practice, we select the weight values empirically according to experimental results, \emph{i.e.}, we set $\lambda_{0}=10$, $\lambda_{1}=1$, and $\lambda_{2}=1$. We found that it is necessary to assign a large weight to the bottom sub-generator since the error back propagation path is relatively long for the bottom layer.

The structure of discriminator in all layers are identical to each other, as in~\cite{shaham2019singan}. Note that those discriminators are independently trained in our approach. 

A big difference between our pyramid and SinGAN~\cite{shaham2019singan} is that SinGAN is trained layer by layer but all sub-generators in our pyramid are \emph{jointly} trained. Actually, we have tried the layer-by-layer training scheme for our model, \emph{e.g.}, we first train $G_{0}$ and then train $G_{1}$ while fixing the parameters of $G_{0}$. Experimental results show that it is relatively difficult for $G_{1}$ to stably achieve convergence with this layer-by-layer training scheme. We will try to find out the underlying reason of such phenomena in future work.


\section{Experiments}
We conduct experiments on Places2 \cite{zhou2015places2}, CelebA-HQ \cite{karras2017progressive}, and DIV2K \cite{agustsson2017ntire} to evaluate our approach. The official train+val splits of Places2 are used to train our object-inpainting model, where each image is randomly cropped into 512$\times$512 resolution. The $28,000$ images in CelebA-HQ are used to train our face-inpainting model, where all images are resized into size 512$\times$512 as in~\cite{lee2020maskgan}. 

Although our two inpainting models are both trained on 512$\times$512 images, they can be used to restore an image in any resolution. 
Besides evaluating our approach on 512$\times$512 testing images, high-resolution testing image are also involved into the evaluation. 
Particularly, we choose DIV2K val as our high-resolution testing dataset, which consists of $100$ images from the Internet in 2k resolution. These images have diverse contents in nature and are suitable for evaluations. Note that the images of DIV2K are randomly cropped into size 512$\times$512 and 1024$\times$1024 for evaluation, since it is very memory-consuming to conduct image inpainting on 2k resolution.

We adopt the original mask generation algorithm in \cite{yu2019free}, which generates center square mask plus random free-form masks for each training sample. All of our experiments are trained with TensorFlow v1.15, CUDA v10.0. The final model needs 5 days to converge on a single NVIDIA Tesla V100 GPU with batch size of 4, and we apply Adam as our optimizer. We are going to release our code in the future.

\begin{figure}[htbp]\vspace{-0.1cm}
    \centering
    \subfigure[Masked]{
    \begin{minipage}[b]{0.18\linewidth}
    \includegraphics[width=1\linewidth]{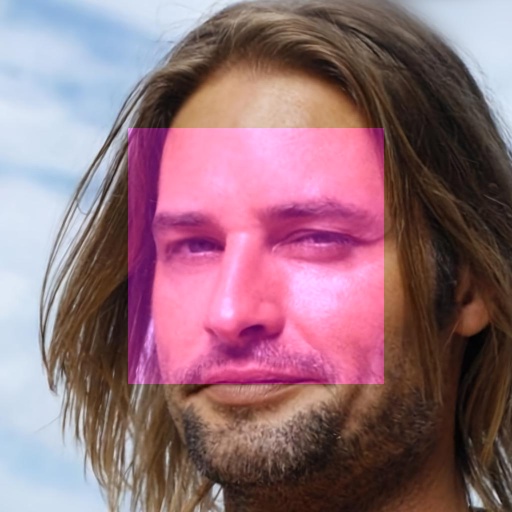}\vspace{1pt}
    \includegraphics[width=1\linewidth]{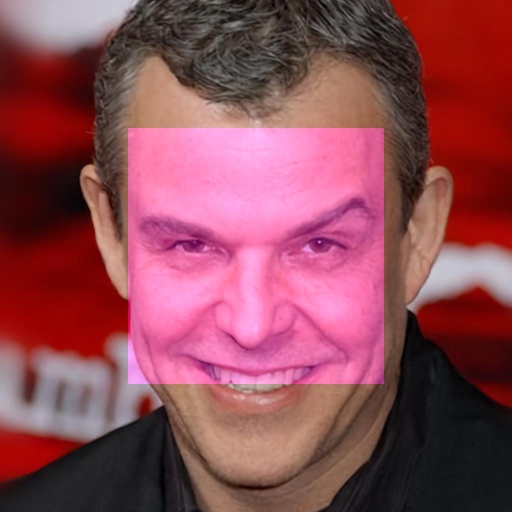}\vspace{1pt}
    \includegraphics[width=1\linewidth]{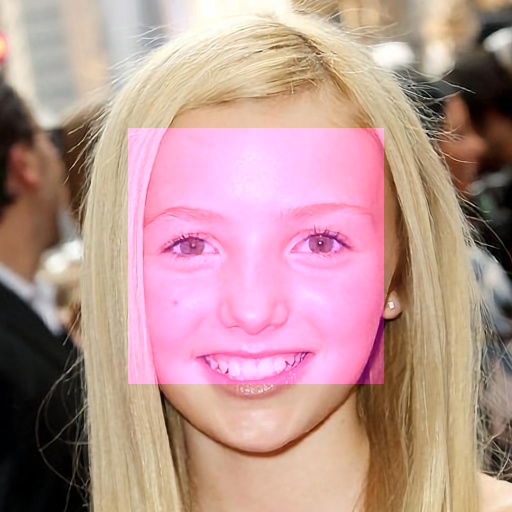}\vspace{1pt}
    \end{minipage}}
    \subfigure[G\&L]{
    \begin{minipage}[b]{0.18\linewidth}
    \includegraphics[width=1\linewidth]{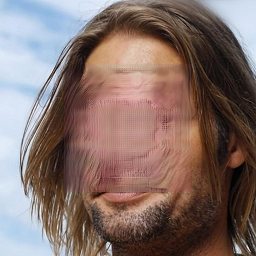}\vspace{1pt}
    \includegraphics[width=1\linewidth]{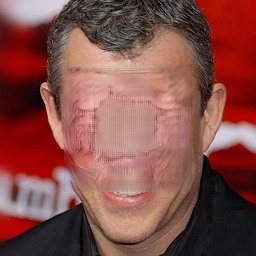}\vspace{1pt}
    \includegraphics[width=1\linewidth]{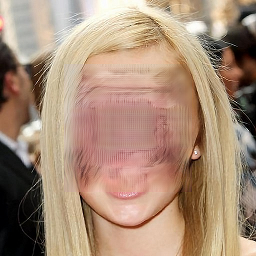}\vspace{1pt}
    \end{minipage}}
    \subfigure[DeepFill]{
    \begin{minipage}[b]{0.18\linewidth}
    \includegraphics[width=1\linewidth]{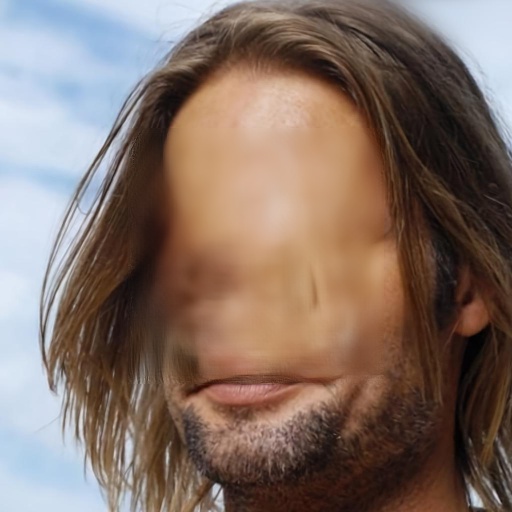}\vspace{1pt}
    \includegraphics[width=1\linewidth]{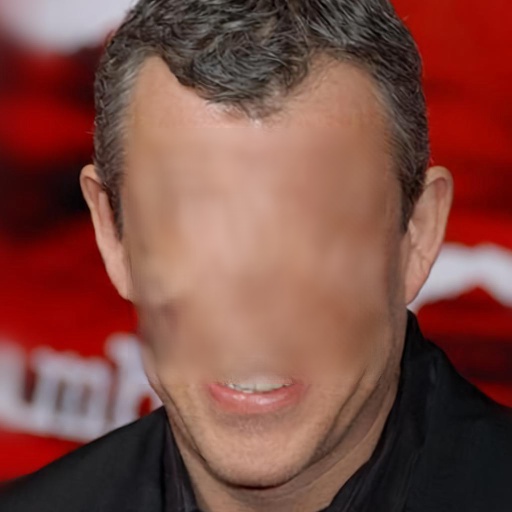}\vspace{1pt}
    \includegraphics[width=1\linewidth]{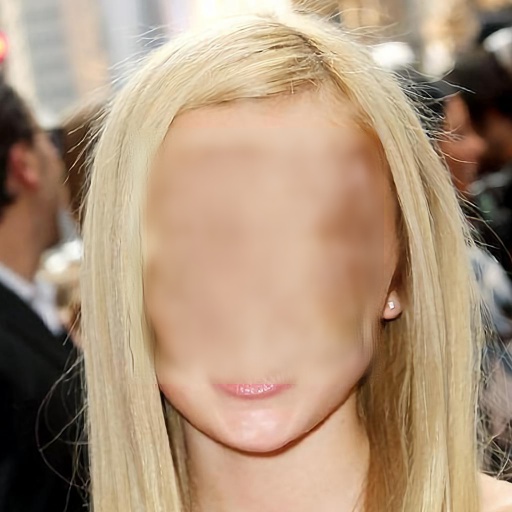}\vspace{1pt}
    \end{minipage}}
    \subfigure[DeepFill$^\dagger$]{
    \begin{minipage}[b]{0.18\linewidth}
    \includegraphics[width=1\linewidth]{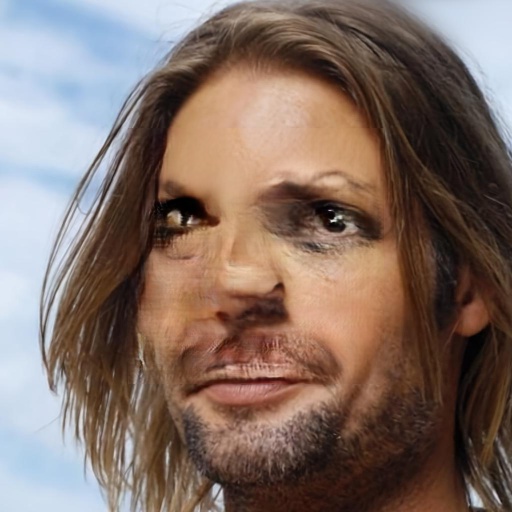}\vspace{1pt}
    \includegraphics[width=1\linewidth]{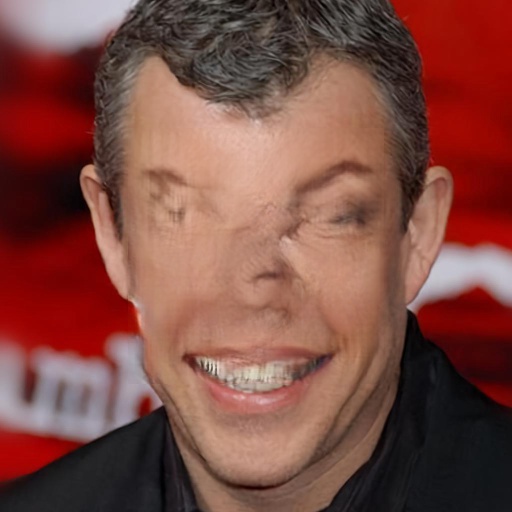}\vspace{1pt}
    \includegraphics[width=1\linewidth]{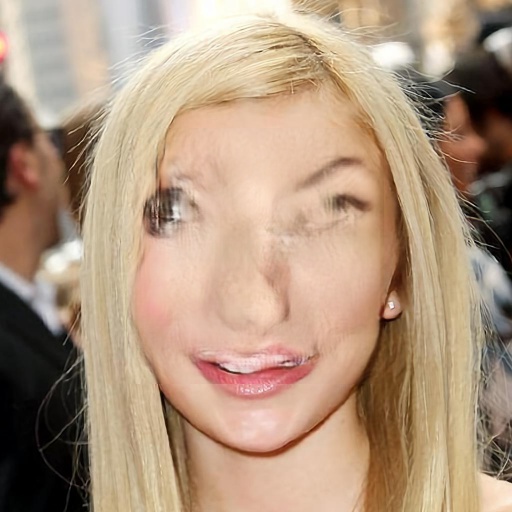}\vspace{1pt}
    \end{minipage}}
    \subfigure[Ours]{
    \begin{minipage}[b]{0.18\linewidth}
    \includegraphics[width=1\linewidth]{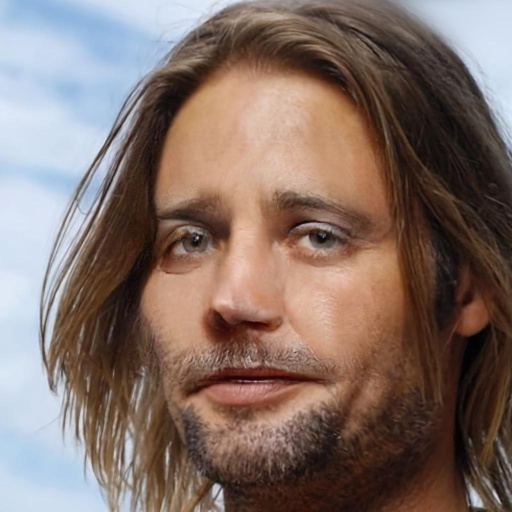}\vspace{1pt}
    \includegraphics[width=1\linewidth]{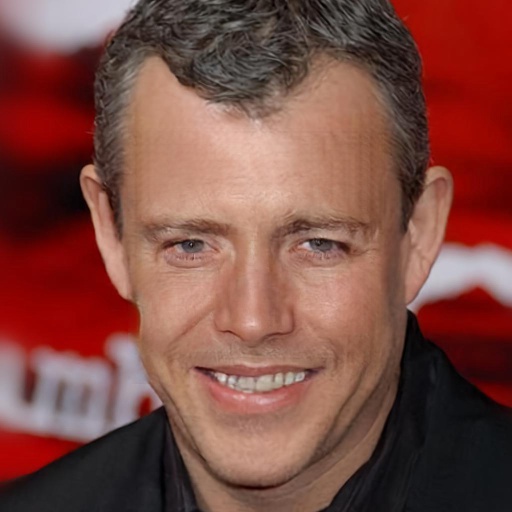}\vspace{1pt}
    \includegraphics[width=1\linewidth]{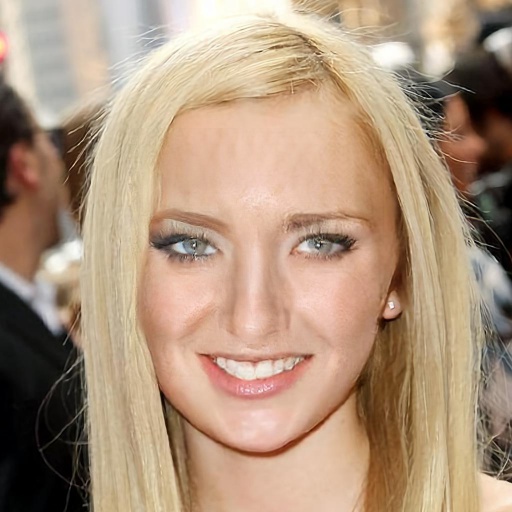}\vspace{1pt}
    \end{minipage}}
    \caption{Qualitative comparisons on CelebA val set (with image resolution 512$\times$512).}
    \label{fig:visual_face}
\end{figure}

\subsection{Comparison to SOTA Methods}
In this section, we compare our model with methods Global\&Local (G\&L) \cite{iizuka2017globally}, MNPS \cite{HR17}, EdgeConnect (EC) \cite{nazeri2019edgeconnect}, DeepFillv2 \cite{yu2019free}, PEN \cite{zeng2019learning}, and HiFill \cite{HiFill}. The original Global\&Local, MNPS, EdgeConnect and PEN are trained with 256$\times$256 images, while HiFill is trained with 512$\times$512 images. For more comparison, we also re-trained DeepFillv2 on 512$\times$512 images, noted as DeepFillv2$^\dagger$.

For numerical comparisons, we evaluate those methods on $L1$ loss, structural similarity index measure (SSIM) \cite{wang2004image}, and peak signal-to-noise ratio (PSNR). To calculate the inference time per image, we test all the images in Places2 validation set with center masks on single NVIDIA GTX 2080Ti GPU.

\textbf{Comparisons on 512$\times$512 images}
Table~\ref{metric} is the comparison of those methods on the datasets Places2 and DIV2K (with images in 512$\times$512 resolution). We could find that our approach outperforms all the other methods on both the free-form mask inpainting and the center mask inpainting.

\begin{figure}[htbp]
    \centering
    \subfigure[Masked]{
    \begin{minipage}[b]{0.23\linewidth}
    \includegraphics[width=1\linewidth]{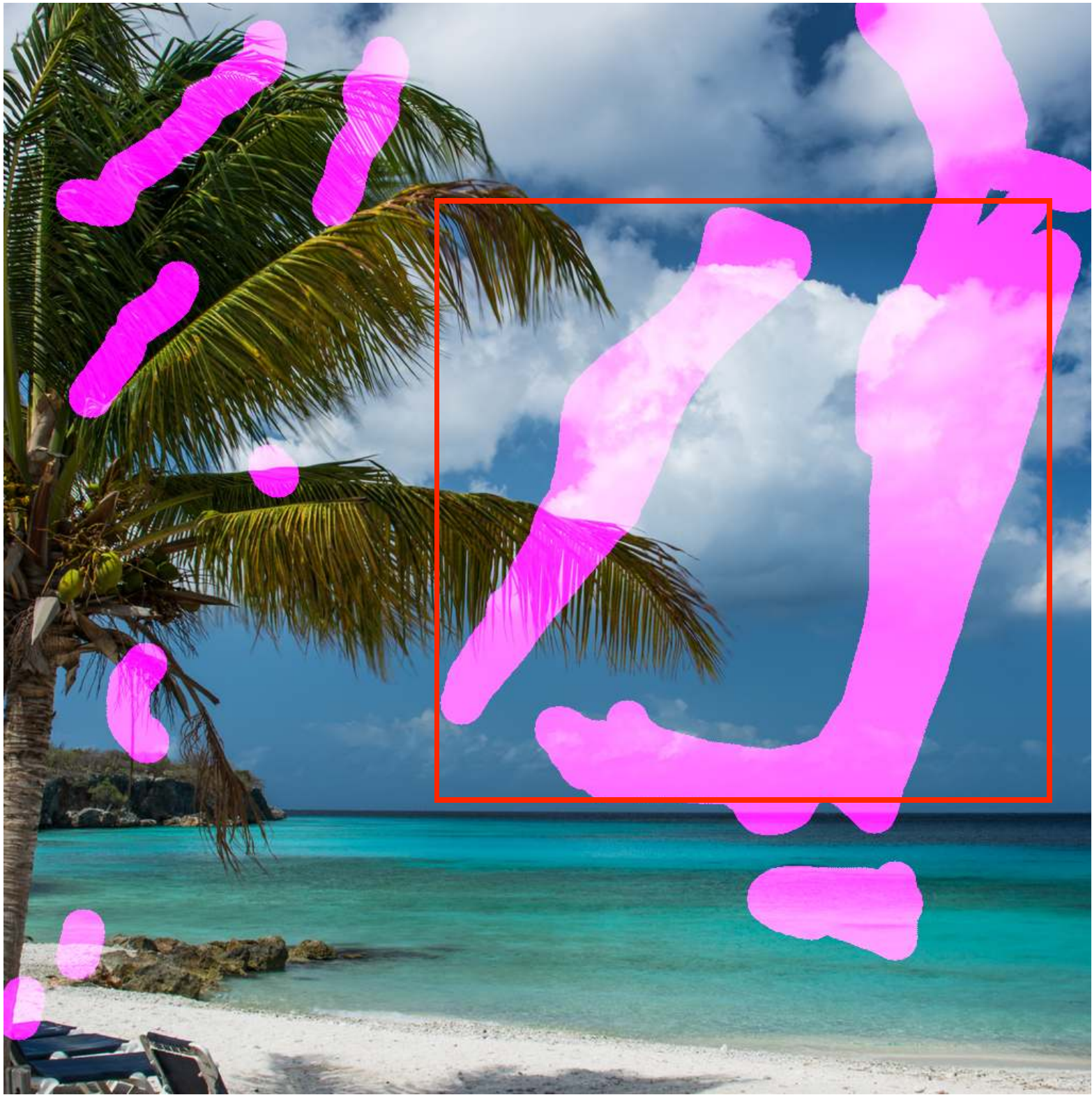}\vspace{1pt}
    \includegraphics[width=1\linewidth]{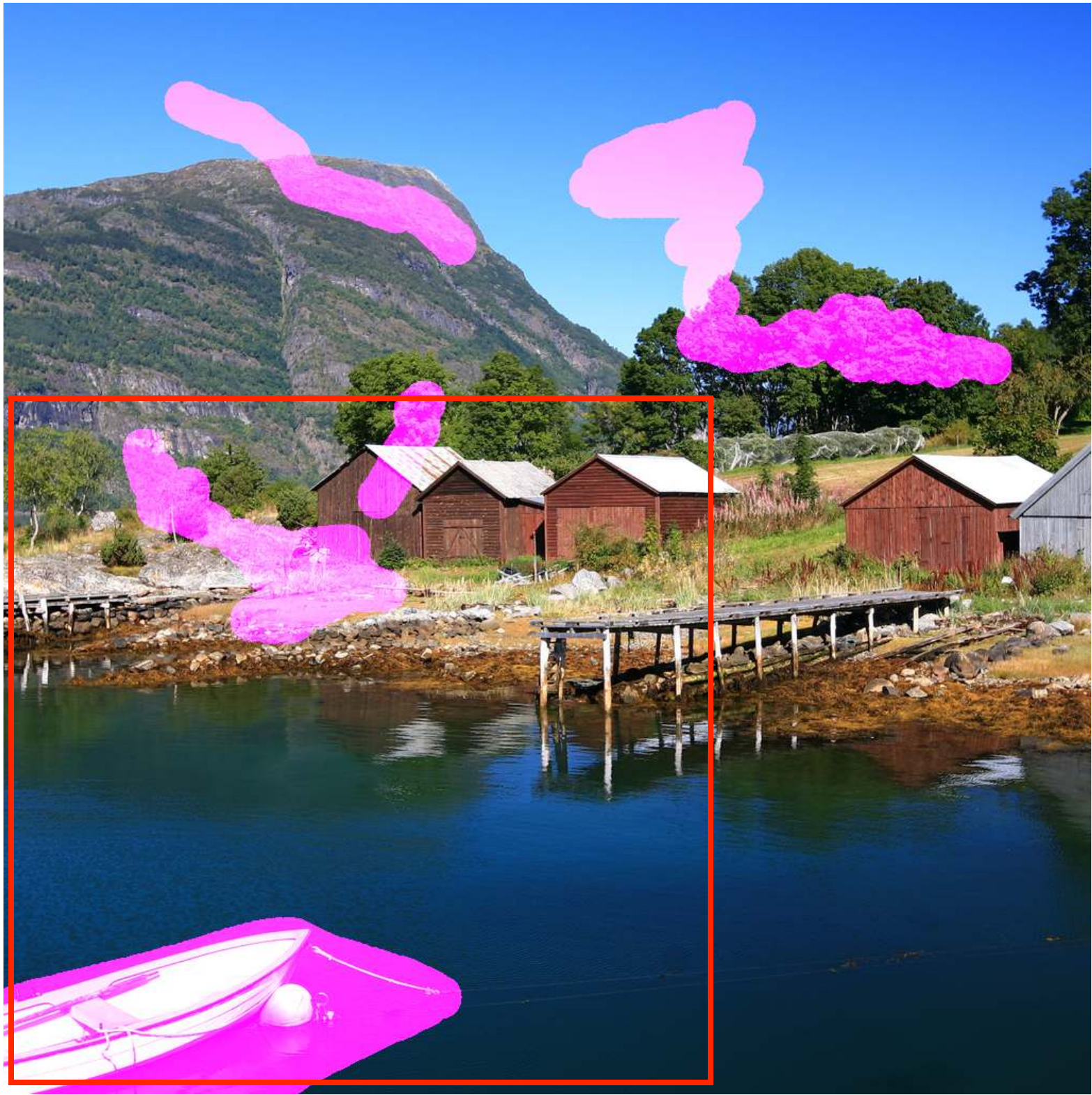}\vspace{1pt}
    \includegraphics[width=1\linewidth]{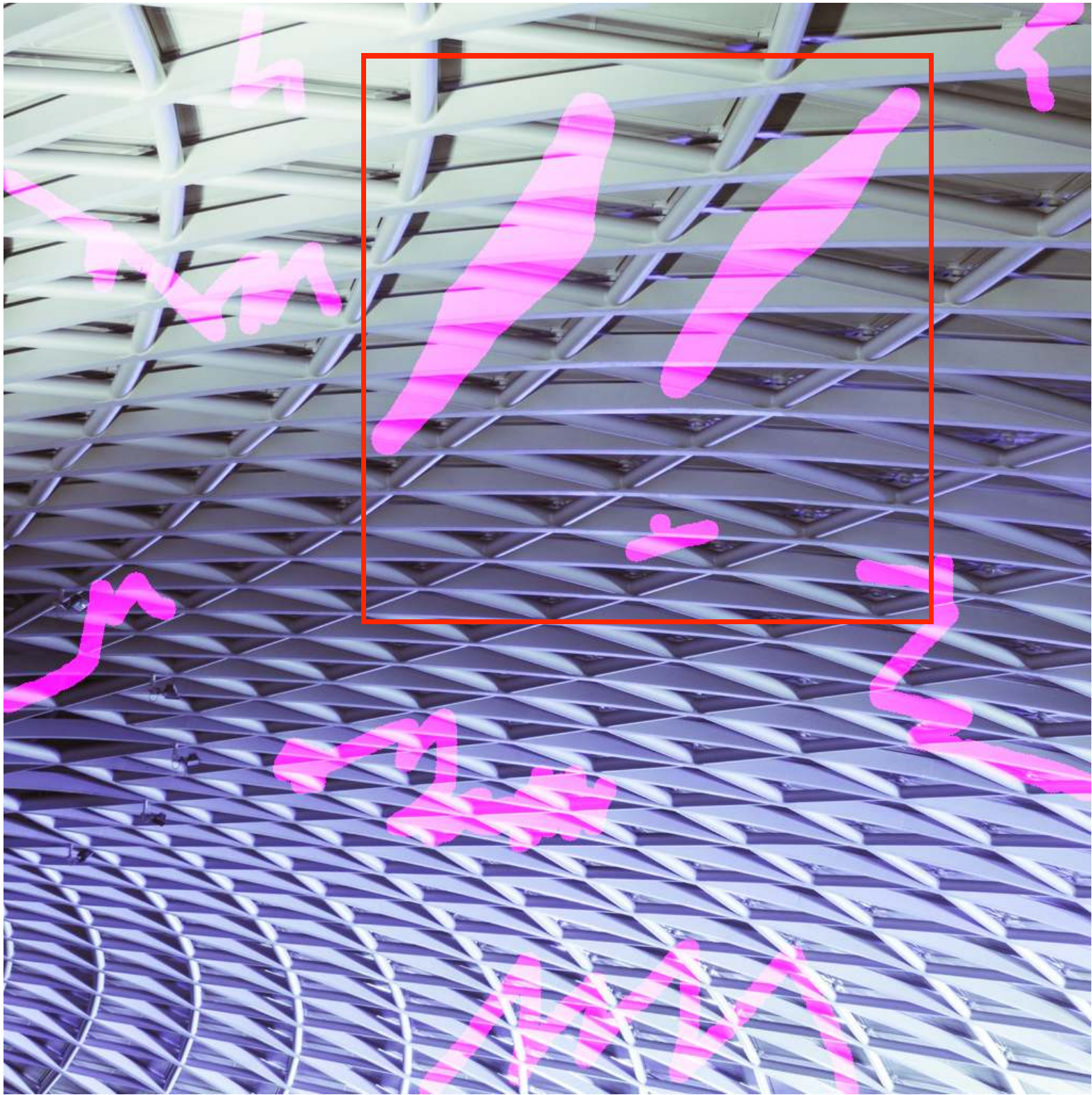}\vspace{1pt}
    \end{minipage}}
    \subfigure[DeepFillv2$^\dagger$]{
    \begin{minipage}[b]{0.23\linewidth}
    \includegraphics[width=1\linewidth]{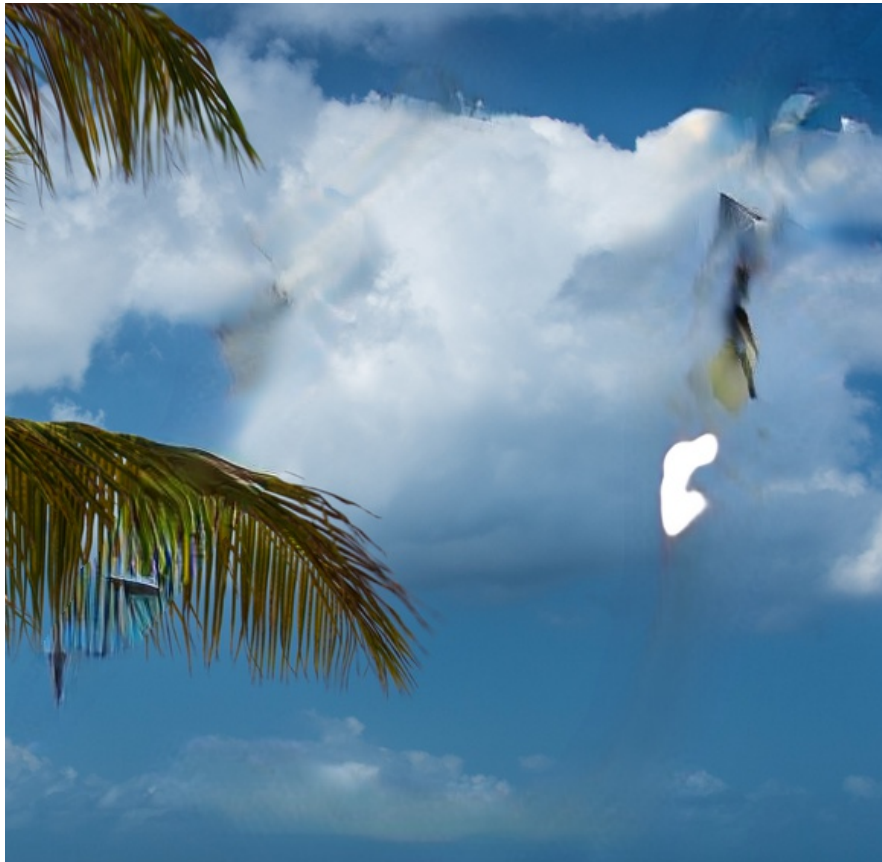}\vspace{1pt}
    \includegraphics[width=1\linewidth]{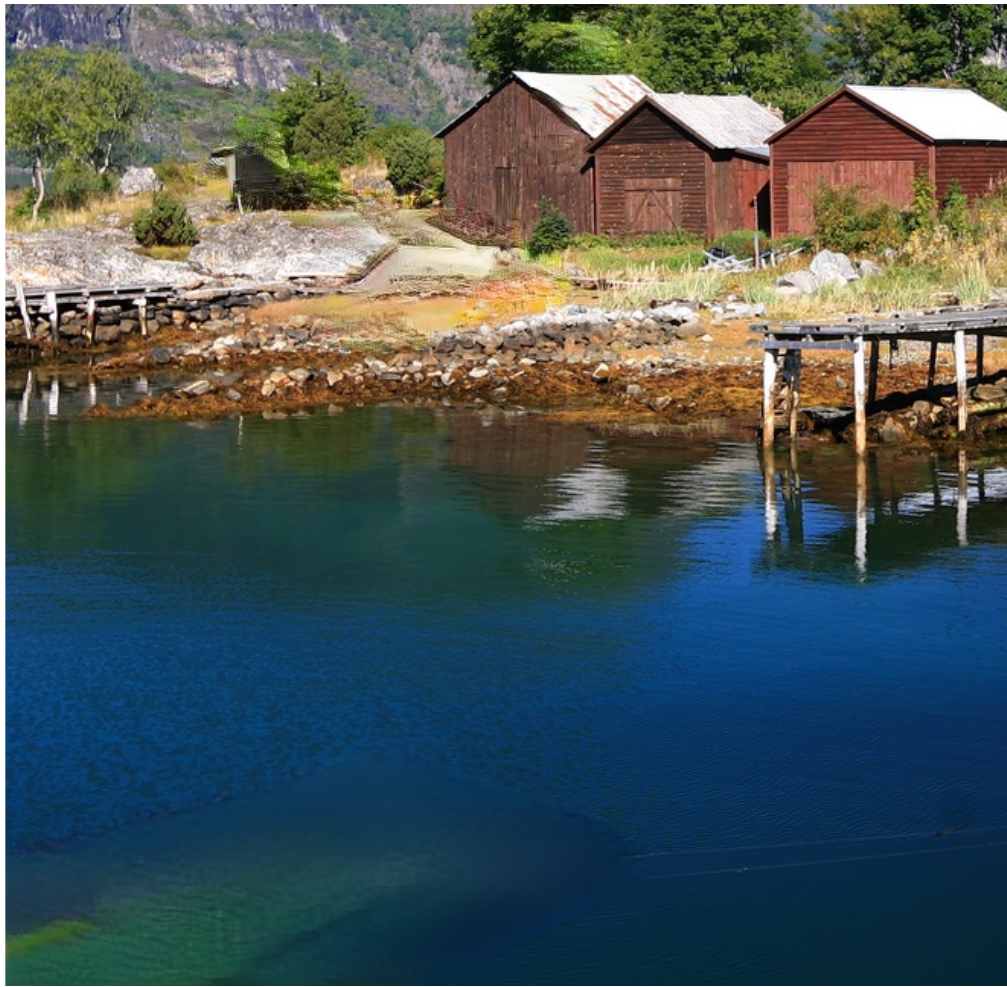}\vspace{1pt}
    \includegraphics[width=1\linewidth]{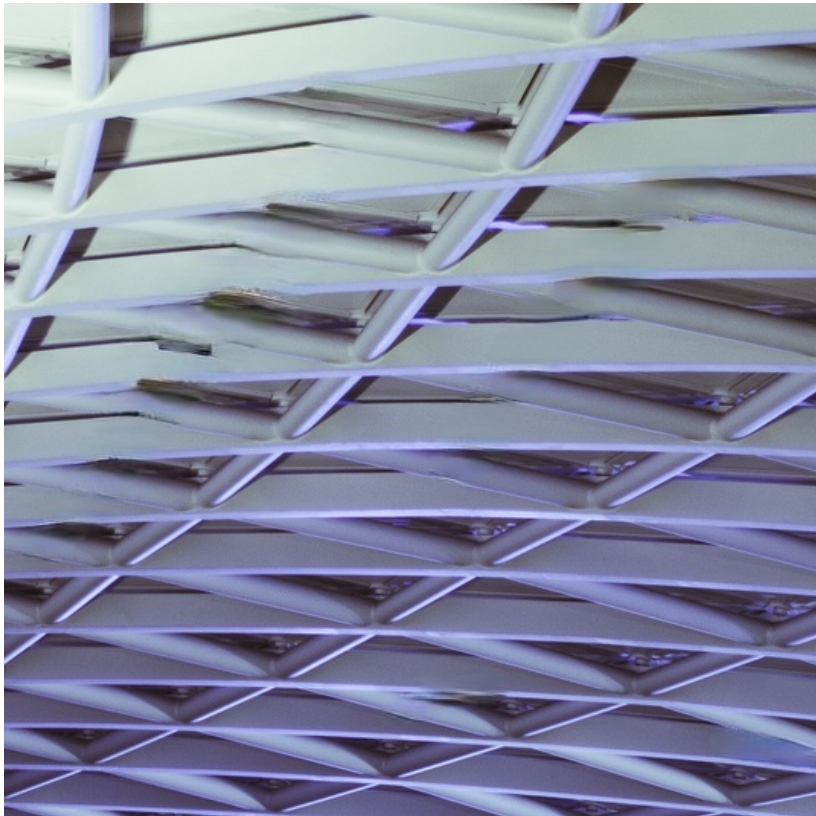}\vspace{1pt}
    \end{minipage}}
    \subfigure[HiFill]{
    \begin{minipage}[b]{0.23\linewidth}
    \includegraphics[width=1\linewidth]{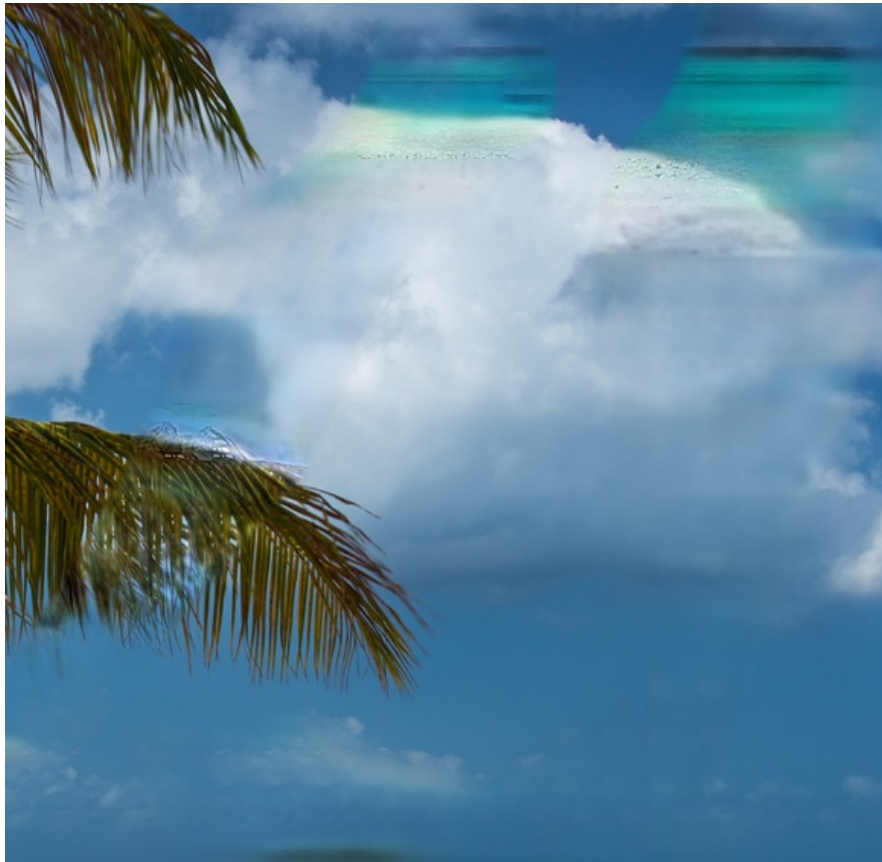}\vspace{1pt}
    \includegraphics[width=1\linewidth]{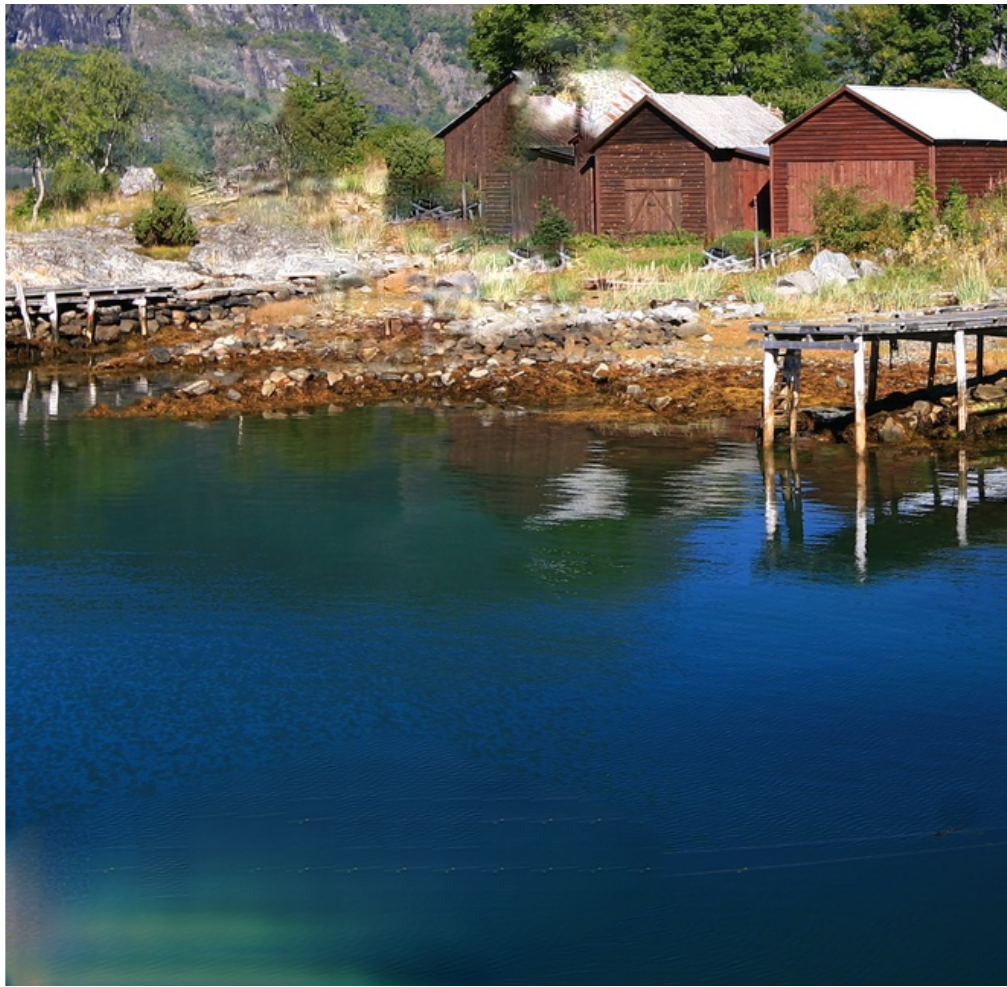}\vspace{1pt}
    \includegraphics[width=1\linewidth]{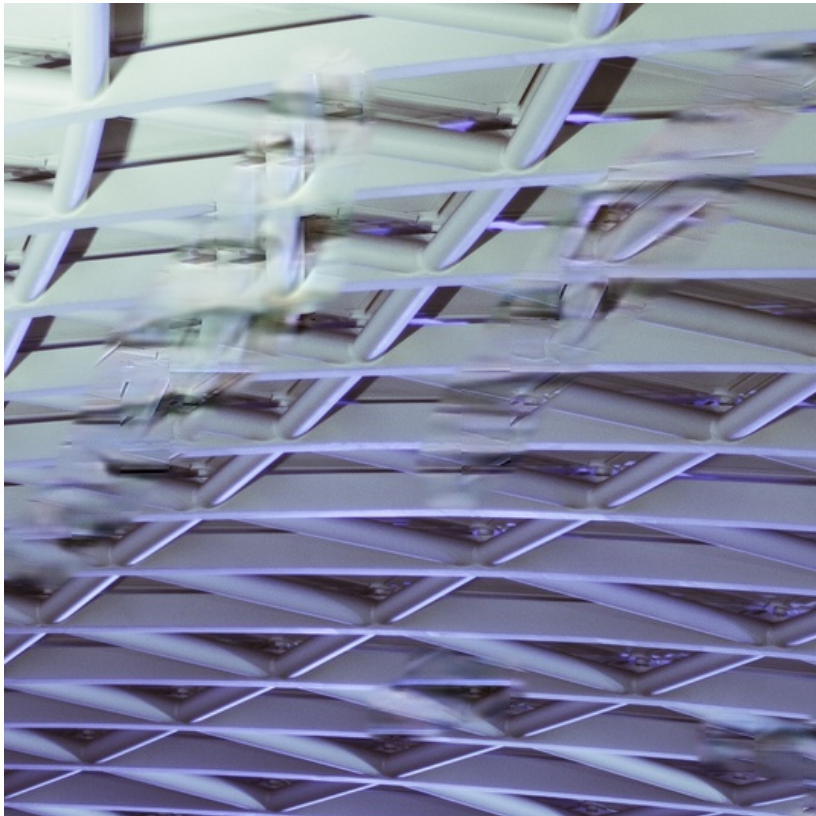}\vspace{1pt}
    \end{minipage}}
    \subfigure[Ours]{
    \begin{minipage}[b]{0.23\linewidth}
    \includegraphics[width=1\linewidth]{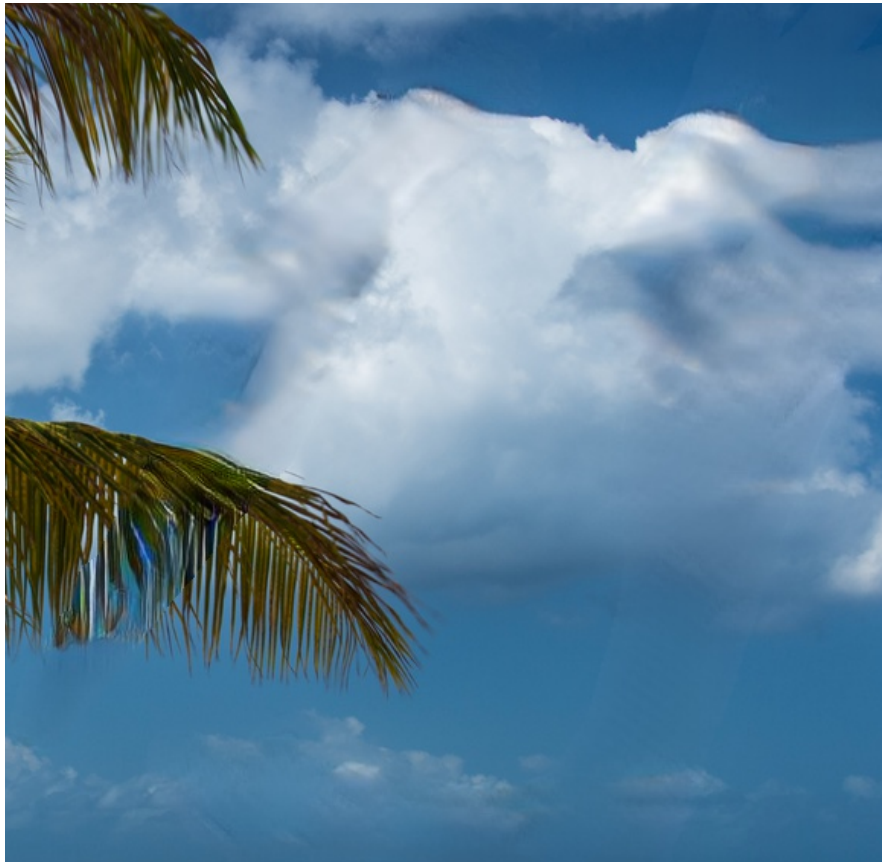}\vspace{1pt}
    \includegraphics[width=1\linewidth]{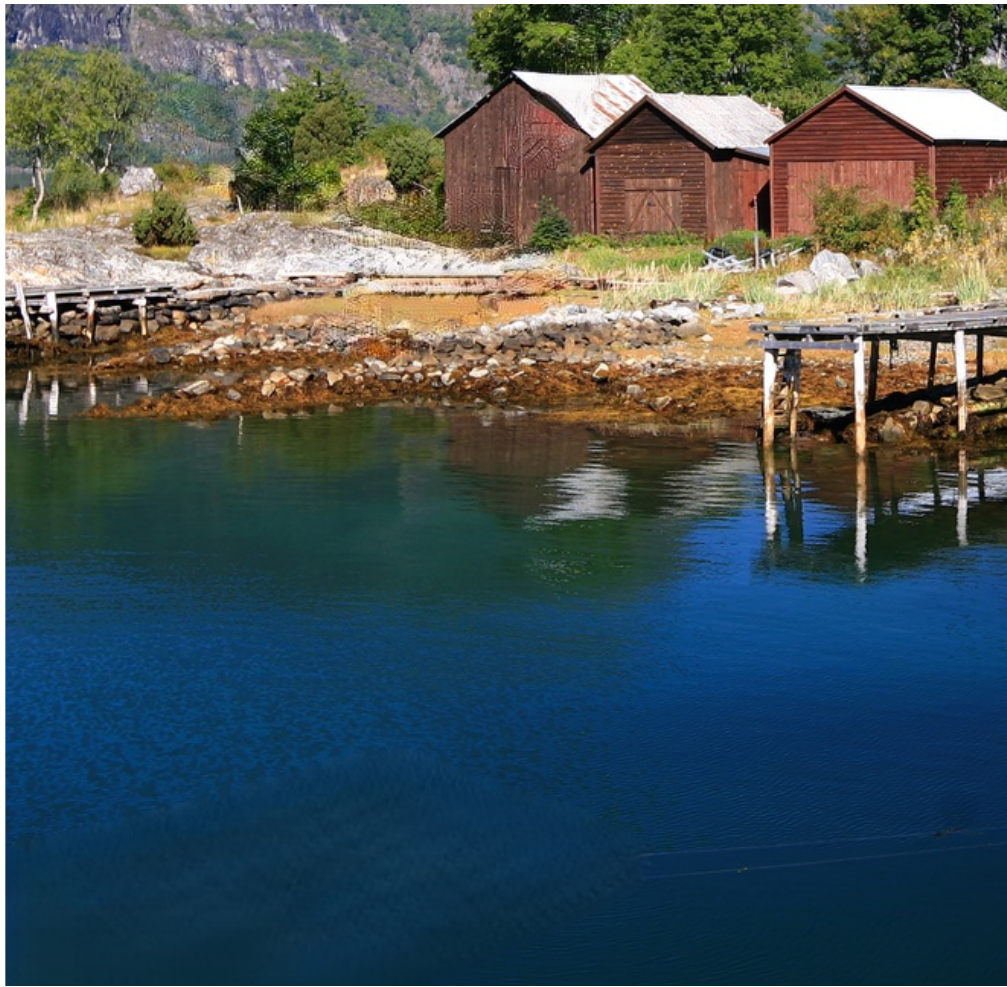}\vspace{1pt}
    \includegraphics[width=1\linewidth]{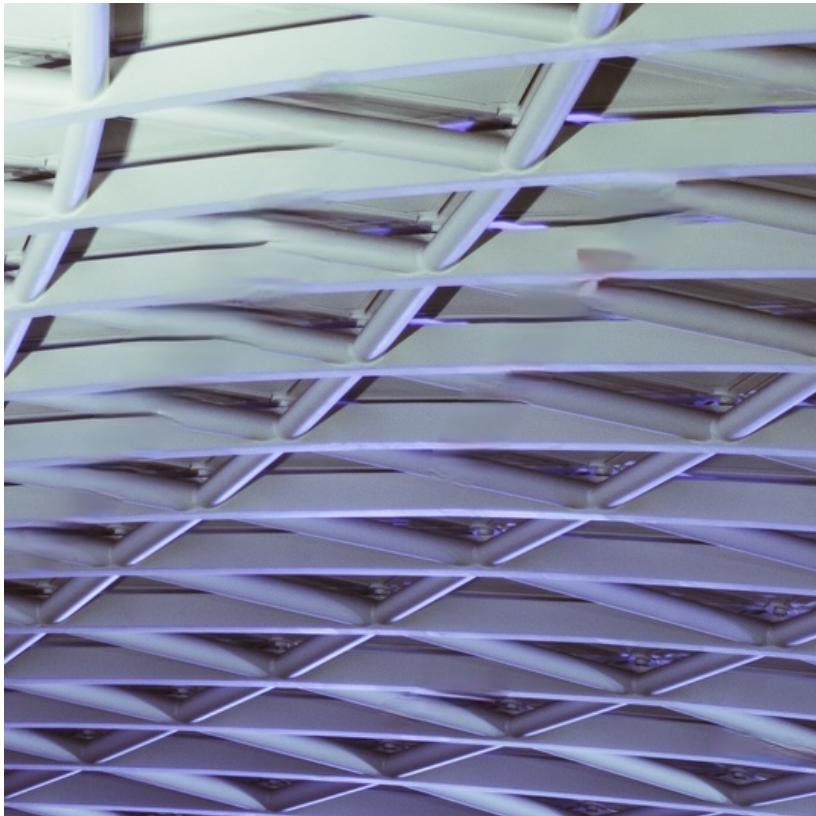}\vspace{1pt}
    \end{minipage}}
    \caption{Qualitative comparisons on DIV2K val set (with image resolution 1024$\times$1024).}
    \label{fig:visual_1024}
\end{figure}

\begin{table}[htbp]
\scriptsize
\centering
\caption{Quantitative comparisons on 1024$\times$1024 images from DIV2K val set. Both center masks and free-form masks setting are considered.}
\begin{tabular}{c|ccc|ccc}
\hline
\multirow{2}*{Method}&\multicolumn{3}{c|}{square mask}& \multicolumn{3}{c}{free-form mask}\\
&SSIM$\uparrow$&PSNR$\uparrow$&L1$\downarrow$&SSIM$\uparrow$&PSNR$\uparrow$&L1$\downarrow$\\
\hline
DeepFill$^\dagger$~\cite{yu2019free} & 0.782& 19.83& 0.052& 0.734& 19.16& 0.072\\
HiFill~\cite{HiFill} & 0.765& 19.58& 0.053&0.700& 18.43& 0.079\\
\hline
\textbf{ours}& \textbf{0.788}& \textbf{20.37}& \textbf{0.049}& 0.734& \textbf{19.36}& \textbf{0.069}\\
\end{tabular}
\label{metric_1024}
\end{table}

Figure~\ref{fig:visual} shows the visual comparison for examples of resolution 512$\times$512 from dataset Places2 and DIV2K. From the results, we could find Global\&Local and EdgeConnect prone to producing color inconsistent and blurry content. The original DeepFillv2 could better recover content on small holes, but it tends to produce artifacts when inpainting large holes. Moreover, PEN is not good at generating realistic high-frequency details. Probably because they are trained on 256$\times$256 images, they cannot properly handle 512$\times$512 inpainting task. 

On the other hand, DeepFillv2$^\dagger$, HiFill and our model are all trained on 512$\times$512 images, and there is an obvious performance improvement. However, the restored content from DeepFillv2$^\dagger$ is not very coherent to their surrounding pixels, and HiFIll is not good at recovering image global structures. In contrast, our approach can not only recover image global structures but also produce coherent image details.

The results on CelebA-HQ are shown in Figure~\ref{fig:visual_face}. Compared to the vanilla DeepFillv2, DeepFillv2$^\dagger$ could properly synthesize some face parts (\emph{e.g.}, eyes).  Nevertheless, it still cannot model the global structures of a face.

\textbf{Comparisons on 1024$\times$1024 images} 
Table~\ref{metric_1024} illustrates the comparison of high-resolution image inpainting (with images resolution 1024$\times$1024). On both free-form mask and center mask, our approach outperforms DeepFillv2$^\dagger$ and HiFill.

\begin{figure}[htbp]\vspace{-0.3cm}
    \centering
    \subfigure[L1]{
    \begin{minipage}[b]{0.475\linewidth}
    \includegraphics[width=1\linewidth]{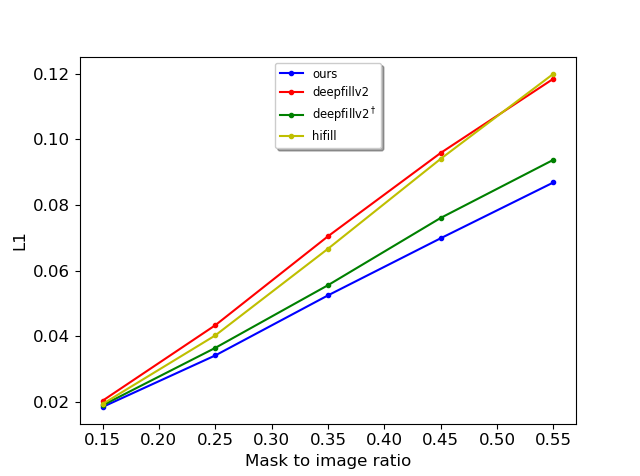}\vspace{1pt}
    \end{minipage}}
    \subfigure[PSNR]{
    \begin{minipage}[b]{0.475\linewidth}
    \includegraphics[width=1\linewidth]{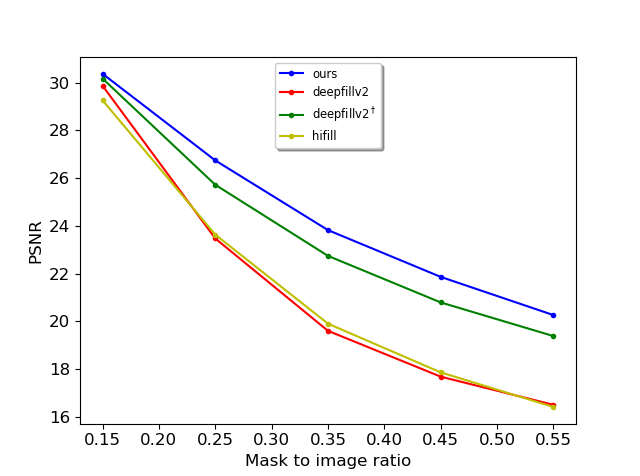}\vspace{1pt}
    \end{minipage}}
    \caption{The performance degradation with respect to the increasing of hole size. With the ratio of hole size increased from $15\%$ to $55\%$, the degradation magnitude of our model is much less than that of DeepFillv2~\cite{yu2019free}. All the experiments are on DIV2K val set with square shape masks.}
    \label{fig:hole}\vspace{-0.3cm}
\end{figure}

The results of these methods on high-resolution image inpainting are shown in Figure~\ref{fig:visual_1024}. Obviously, it is much more challenging to generate coherent and realistic image details on 1024$\times$1024 images. For example, DeepFillv2$^\dagger$ tends to generate incorrect semantic content to fill in the hole (e.g., the missing `cloud' region are filled with `tree' pixels), and there is visible color discrepancy for HiFill inpainting results. At the same time, our approach's inpainting results look more relevant and realistic.

\subsection{Large-hole Inpainting}

It is known that when the corrupted area is large the image inpainting becomes much more challenging. As shown in Figure~\ref{fig:visual}, when the hole size becomes large, the center of the hole tends to have color discrepancy and blurriness. 

Recent work found that such issues could be alleviated by adopting the progressive learning strategy~\cite{Soo2020Zoom}, without the needs of modifying the generative model itself. The progressive learning strategy is to divide the training procedure into several stages so that the small-hole training data are first used to train the model at early stages whereas large-hole training data will be used to fine-tune the model at later stages. Even if it is effective sometimes, such strategy still has some limitations. For example, it is hard to decide when to move from one stage to the next stage.

As aforementioned, we have illustrated that our pyramid generator could not only align to the progressive learning strategy but also avoid the need for explicitly separating the training procedure into several stages. In this section, we will evaluate our method by gradually increasing the hole size and show its advantage on large hole inpainting.

Specifically, the center mask setting is adopted in this experiment, since it is easy to control the ratio of hole size to the image size. We gradually increase such ratio from $15\%$ to $55\%$ with interval of $10\%$. As shown in Figure~\ref{fig:hole}, the red line is the performance of DeepFillv2~\cite{yu2019free} and the blue line indicates our model performance. There is a clear performance degradation for both methods when the ratio increases, but the degradation magnitude of our model is much less than that of DeepFillv2.

\begin{figure}[htbp]\vspace{-0.3cm}
    \centering
    \subfigure[L1]{
    \begin{minipage}[b]{0.475\linewidth}
    \includegraphics[width=1\linewidth]{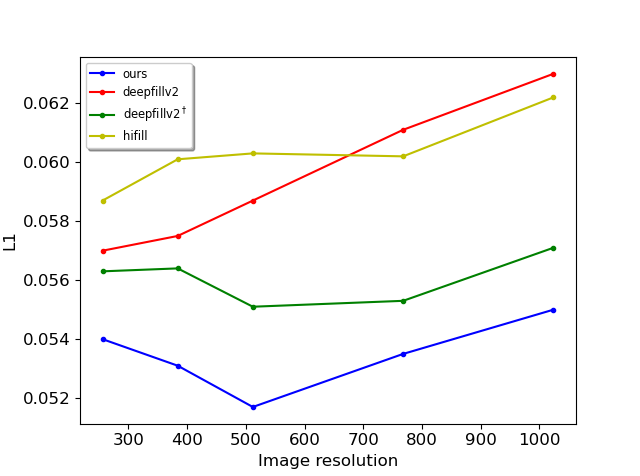}\vspace{1pt}
    \end{minipage}}
    \subfigure[PSNR]{
    \begin{minipage}[b]{0.475\linewidth}
    \includegraphics[width=1\linewidth]{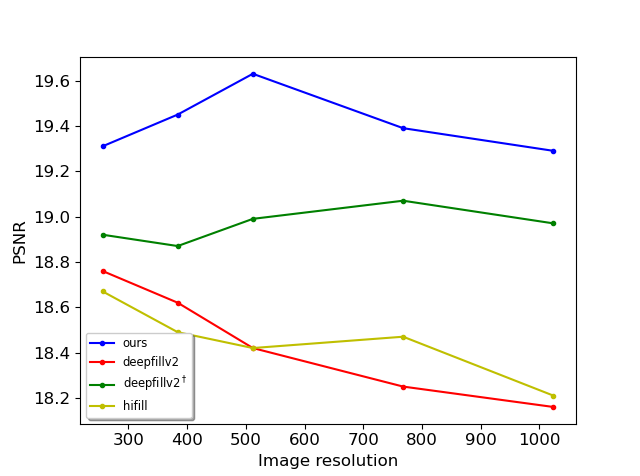}\vspace{1pt}
    \end{minipage}}
    \caption{The performance degradation with respect to the increasing of image resolution. With the image resolution increased from $256$ to $1024$, our model could better cope against the performance degradation, and hence our model is more suitable for high-resolution image inpainting. Experiments are conduct on DIV2K val set with center square masks.}
    \label{fig:reso}\vspace{-0.3cm}
\end{figure}

Besides, it is obvious that increasing resolution of training images also benefits large-hole inpainting task (\emph{e.g.}, DeepFillv2$^\dagger$ outperforms DeepFillv2), but it is still inferior to our approach.

\subsection{High-resolution Image Inpainting}\label{section:reso}
With the rapid development of high quality camera devices, the images we meet in daily life are often high-resolution. Thus, the real-world inpainting task is often about high-resolution image inpainting. 

On the other hand, many previous works have demonstrated that increasing the resolution of training images is critical to the inpainting performance, especially for the high-resolution image inpainting task \cite{zeng2019learning,HiFill}.

In this section, we conduct some experiments to study such observations. Specifically, we train the same model DeepFillv2~\cite{yu2019free} with images of different resolution: the original DeepFillv2 is trained with images of $256\times 256$ resolution, while the model DeepFillv2$^\dagger$ is trained with images of $512\times 512$ resolution. From Figure~\ref{fig:reso}, we can see that the inpainting performance could be improved by increasing the resolution of training images, which validates the conclusion of previous work.

Besides, Figure~\ref{fig:reso} tells us that the inpainting performance will decrease when the resolution of testing image is increasing. This indicates that the high-resolution image inpainting task is much more difficult than low-resolution inpainting task. Moreover, it is worth noting that such performance degradation is different between the model DeepFillv2 and DeepFillv2$^\dagger$: the DeepFillv2$^\dagger$ could better cope against the performance degradation than the DeepFillv2, \emph{i.e.,} the green line drops slower than the red line according to the metric PSNR. Therefore, it is critical to use high-resolution images as training data. 

More importantly, we also evaluate our pyramid generator (noted as blue line in Figure~\ref{fig:reso}). Compared to the model DeepFillv2$^\dagger$ and HiFill~\cite{HiFill}, we can see that our model outperforms those methods, although they are all trained with images of same resolution (\emph{i.e.}, $512\times 512$). It indicates that our model could fully exploit the benefits of learning with high-resolution images, and hence our model is good at high-resolution image inpainting. 

\subsection{Ablation Study}\label{section:ablation}
Except the recent architecture of our pyramid generator described in Section~\ref{section:Approach}, there are many other design options. We have evaluated those design options and will discuss their performance in this section. Such evaluation are conducted on DIV2K dataset under the same settings as the previous experiments.

\begin{table}[htbp]
\small
\centering
\caption{Comparisons between our three-layer model and the two-layer variants.}
\begin{tabular}{c|ccc}
\hline
~&SSIM$\uparrow$&PSNR$\uparrow$&L1$\downarrow$\\
\hline
$PG_{128+256}$& 0.809& 24.07 & 0.042\\
$PG_{256+512}$& 0.827& 26.03 & 0.035\\
\hline
$PG_{128+256+512}$ (Ours) & \textbf{0.840}& \textbf{26.74} &\textbf{0.034}\\
\end{tabular}
\label{ablation:scale}
\end{table}

\begin{figure}[htbp]\vspace{-0.4cm}
    \centering
    \subfigure[Masked]{\includegraphics[width=0.4\linewidth]{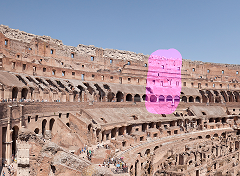}}\vspace{0pt}
    \subfigure[$PG_{128+256}$]{\includegraphics[width=0.4\linewidth]{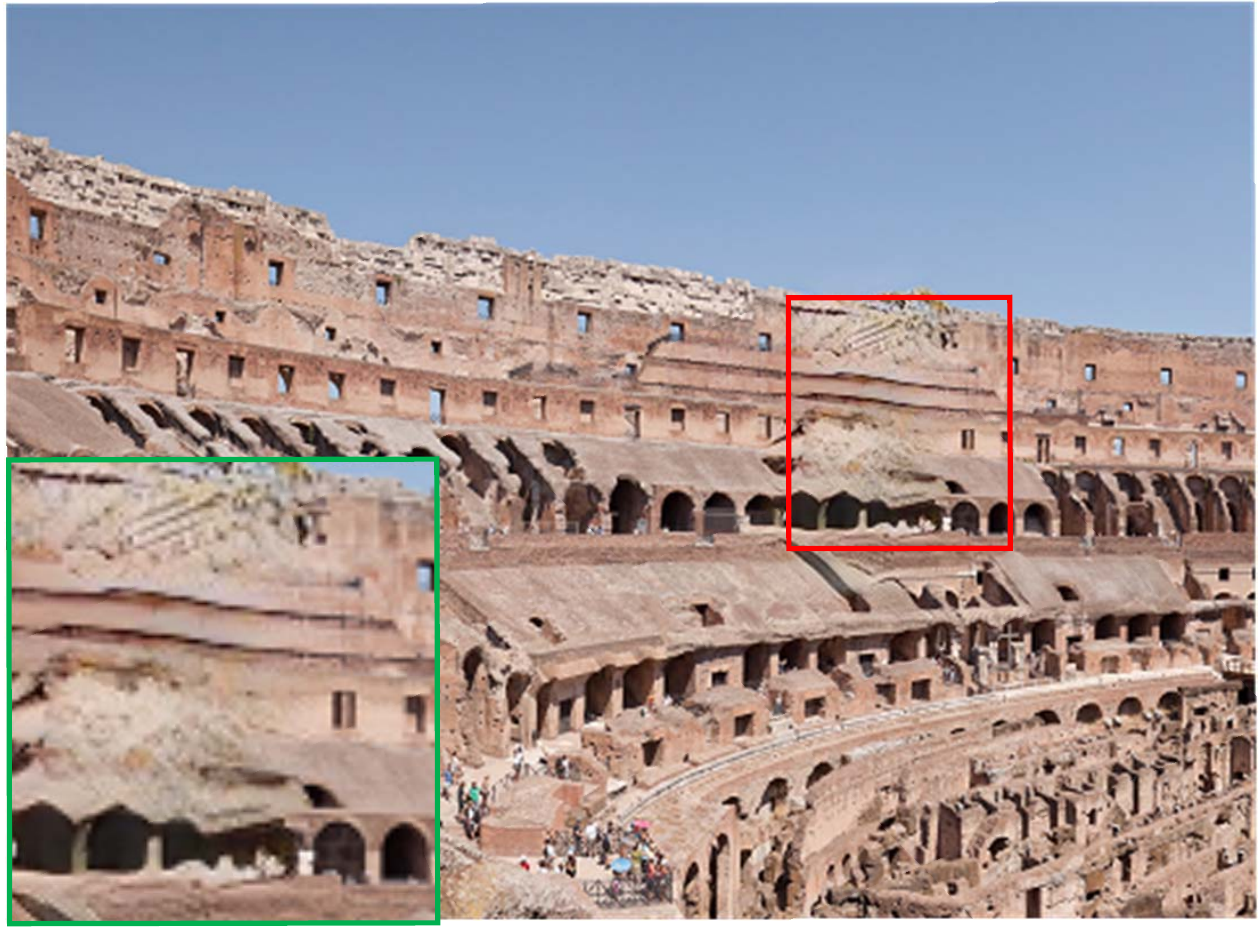}}\vspace{0pt}
    \\
    \subfigure[$PG_{256+512}$]{\includegraphics[width=0.4\linewidth]{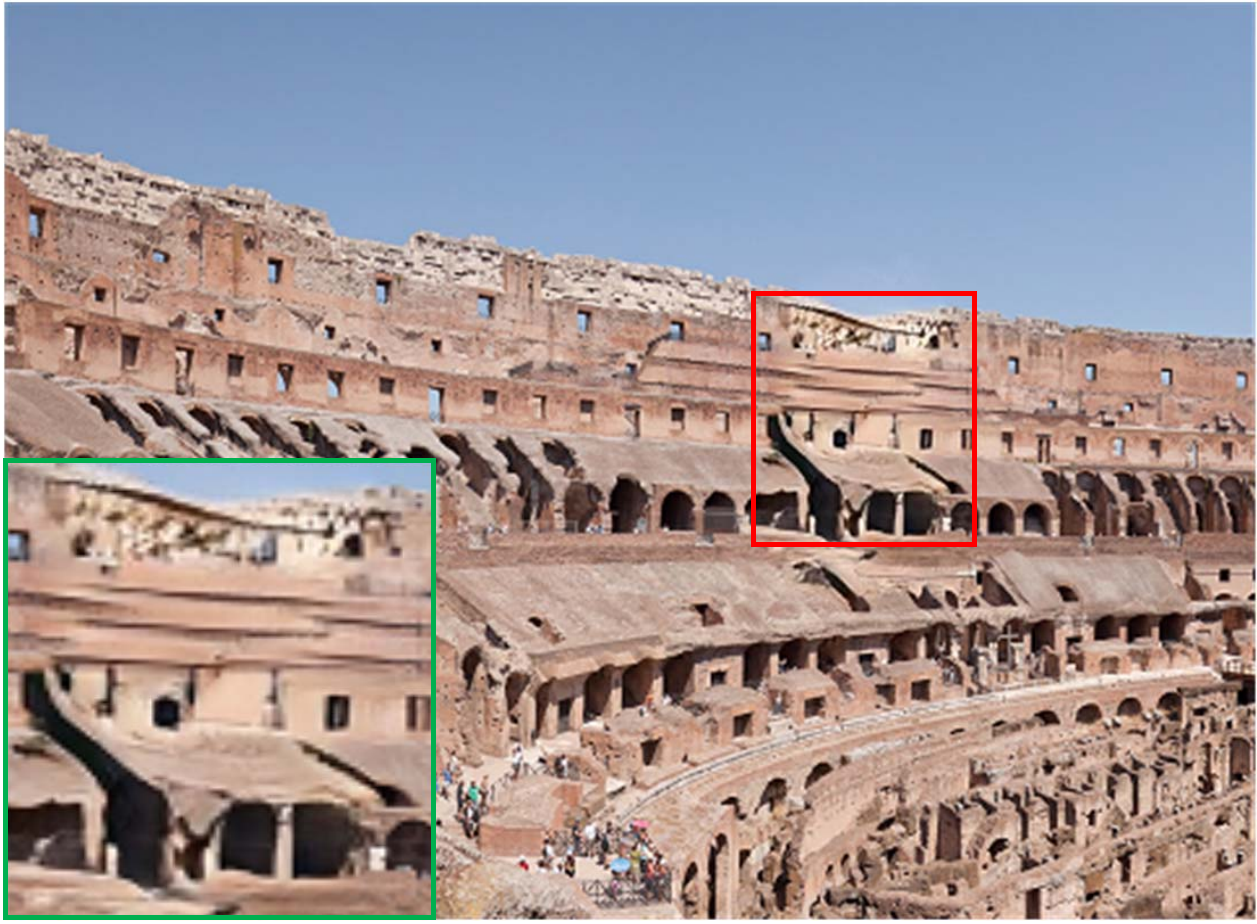}}\vspace{0pt}
    \subfigure[$PG_{128+256+512}$]{\includegraphics[width=0.4\linewidth]{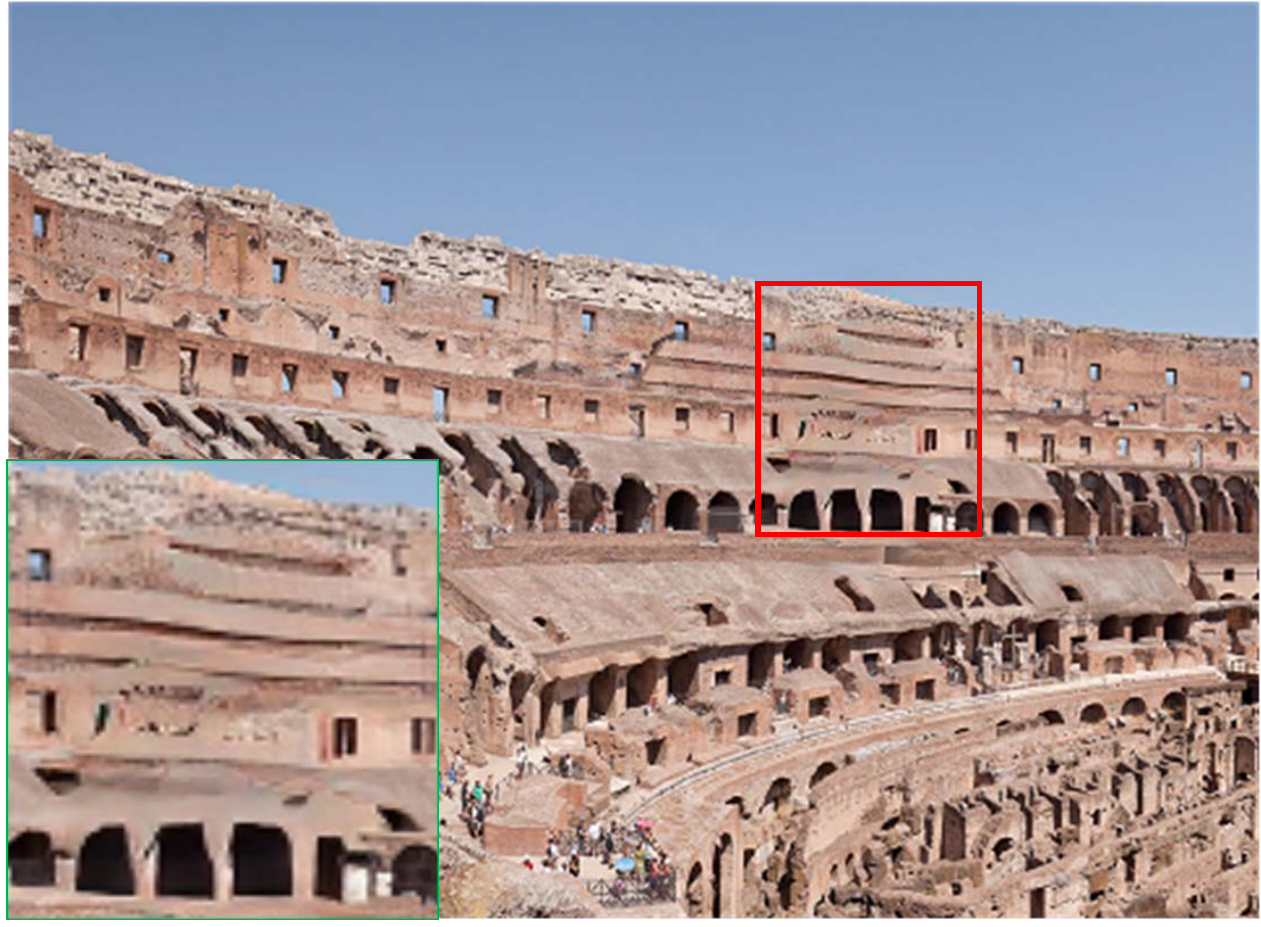}}\vspace{0pt}
    \caption{Qualitative comparisons between our three-layer model and the two-layer variants.}
    \label{fig:ablation-visual}
\end{figure}

\textbf{How many layers should we have?} Since our pyramid generator has a multi-layer architecture where each layer corresponds to a specific image scale, we could have many choices for the number of layers. Apart from our current implementation of 3 layers, we also try two variants of two-layer architecture: one variant is the combination of the 128-layer and 256-layer (noted as $PG_{128+256}$ ), the other variant is the combination of 256-layer and 512-layer (noted as $PG_{256+512}$). 

The comparisons of our recent three-layer model and the two variants are illustrated in Table~\ref{ablation:scale} and Figure~\ref{fig:ablation-visual}. It shows that the two variants have distinct drawbacks. The model $PG_{128+256}$ tends to generate blurry content due to its disadvantage on modeling image high-frequency details. On the other hand, the model $PG_{256+512}$ tends to generate incoherent image structures because of the lacking of image global information. In contrast, our three-layer model $PG_{128+256+512}$ enjoys the advantages of the two variants, and could generate both coherent image structures and realistic image details.

\textbf{How to fuse different layers?} 
As mentioned above, there are many possible fusion options, and we compare them in this section. First, we consider feature-level fusion, which can be further categorized as coarse-stage fusion (\emph{i.e.}, fuse with $G_{1}^{c}$) and refine-stage fusion (\emph{i.e.}, fuse with $G_{1}^{r}$). Next, we consider image-level fusion. Besides the fusion scheme adopted in our model, there are additional two options. Since there are two path for the $G_{1}^{r}$ (\emph{i.e.}, attention path and non-attention path), we can feed the output of $G_{0}$ only to one path. From Table~\ref{ablation:fusion}, we can see that feature-level refine-stage fusion cannot converge, and our fusion scheme is the best among them. 

\begin{table}[htbp]
\footnotesize
\centering
\caption{Quantitative comparisons on different fusion strategies.}
\begin{tabular}{c|cc|ccc}
\hline
&\multicolumn{2}{c|}{feature-level}&\multicolumn{3}{c}{image-level}\\
\cline{2-6}

& coarse & refine & att. path  & non-att path &ours\\
\hline
SSIM$\uparrow$& 0.786&-&0.825&0.833&\textbf{0.840}\\
PSNR$\uparrow$& 23.16&-&25.94&25.96&\textbf{26.74}\\
L1$\downarrow$& 0.044&-&0.036&0.036&\textbf{0.034}

\end{tabular}
\label{ablation:fusion}
\end{table}

\textbf{Is adaptive dilation effective?} Due to the pyramid structure, our approach could adopt the \emph{adaptive dilation} mechanism, which means the sub-generators at different layers could have different number of dilation layers, and different dilation rate configurations. In this section, we use the baseline that all sub-generators have same number of dilation layers and same dilation rate configurations. In particular, it has 4 dilation layers with dilation rates $\{2, 4, 8, 16\}$, which is the same as \cite{yu2019free}. In contrast, in our approach $G_{0}$ has 4 dilation layers with dilation rates $\{2, 4, 8, 12\}$, and both $G_{1}$ and $G_{2}$ have 3 dilation layers with dilation rates $\{2, 4, 8\}$. As shown in Table~\ref{ablation:dilation}, we find that our adaptive dilation mechanism benefits the performance improvement compared to the baseline methods. 

\begin{table}[htbp]
\small
\centering
\caption{Quantitative comparisons between the standard and our adaptive dilation scheme.}
\begin{tabular}{c|ccc}
\hline
~&SSIM$\uparrow$&PSNR$\uparrow$ & L1$\downarrow$\\
\hline
standard dilation& 0.829& 26.38 & 0.035\\
\hline
 adaptive dilation (Ours)& \textbf{0.840}& \textbf{26.74} & \textbf{0.034}\\
\end{tabular}
\label{ablation:dilation}\vspace{-0.4cm}
\end{table}

\section{Conclusion}
This paper proposes a `structure first detail next' workflow for image inpainting. In particular, we introduce a Pyramid Generator by stacking several sub-generators, where image global structures and local details could be better separately modeled at different pyramid layers. Notice that our approach has a progressive learning scheme which allows it to restore images with large masks. In addition, our model is suitable for inpainting high-resolution images. Extensive experiments show that our approach outperforms the other state-of-the-art methods.

{\small
\bibliographystyle{ieee_fullname}
\bibliography{main}
}

\end{document}